%% file: main.tex
\documentclass[sigconf]{acmart}
\input{header}

\AtBeginDocument{%
  \providecommand\BibTeX{{%
    \normalfont B\kern-0.5em{\scshape i\kern-0.25em b}\kern-0.8em\TeX}}}

\begin{document}

\title{Fair Active Learning}

\author{Hadis Anahideh}
\authornotemark[1]
\email{hadis@uic.edu}
\affiliation{\institution{University of Illinois at Chicago}
  \country{USA}
}

\author{Abolfazl Asudeh}
\email{asudeh@uic.edu}
\affiliation{%
  \institution{University of Illinois at Chicago}
    \country{USA}
}

\author{Saravanan Thirumuruganathan}
\email{sthirumuruganathan@hbku.edu.qa}
\affiliation{%
  \institution{QCRI, HBKU}
    \country{Qatar}
}

\begin{abstract}
% \hadis{will revise after finishing the results}
Machine learning (ML) is increasingly being used in high-stakes applications impacting society. Therefore, it is of critical importance that ML models do not propagate discrimination. Collecting accurate labeled data in societal applications is challenging and costly. Active learning is a promising approach to build an accurate classifier by interactively querying an oracle within a labeling budget. We design algorithms for fair active learning that carefully selects data points to be labeled so as to balance model accuracy and fairness.  We demonstrate the effectiveness and efficiency of our proposed algorithms over widely used benchmark datasets using demographic parity and equalized odds notions of fairness.

\end{abstract}

%Sara: commented these for some space. We can uncomment during CRV.
%\begin{CCSXML}
%<ccs2012>
%   <concept>
%       <concept_id>10010147.10010257.10010282.10011304</concept_id>
%       <concept_desc>Computing methodologies~Active learning settings</concept_desc>
%       <concept_significance>500</concept_significance>
%       </concept>
%   <concept>
%       <concept_id>10010405.10010455</concept_id>
%       <concept_desc>Applied computing~Law, social and behavioral sciences</concept_desc>
%       <concept_significance>500</concept_significance>
%       </concept>
%   <concept>
%       <concept_id>10003120.10003130</concept_id>
%       <concept_desc>Human-centered computing~Collaborative and social computing</concept_desc>
%       <concept_significance>300</concept_significance>
%       </concept>
% </ccs2012>
%\end{CCSXML}

%\ccsdesc[500]{Computing methodologies~Active learning settings}
%\ccsdesc[500]{Applied computing~Law, social and behavioral sciences}
%\ccsdesc[300]{Human-centered computing~Collaborative and social computing}

%\keywords{active learning, algorithmic fairness, limited labeled data}

\maketitle
\input{intro}

\input{related_work}
\input{background}
\input{tech/framework}
\input{tech/opt}
\input{tech/fbc}
\input{tech/otherfairness}
% \input{technical}
% \input{fbc}
% \input{otherfairness}
\input{results}
\input{conclusion}
\vspace{-3mm}
\bibliographystyle{ACM-Reference-Format}
\bibliography{ref.bib}
\newpage
\section*{Appendix}
\appendix
\input{tech/fbc-maintain}

\input{tech/proofs}
\input{plots-new-app}

\end{document}

%% file: header.tex
\usepackage{url}            % simple URL typesetting
\usepackage{booktabs}       % professional-quality tables
\usepackage{amsfonts}       % blackboard math symbols
\usepackage{nicefrac}       % compact symbols for 1/2, etc.
\usepackage{microtype}      % microtypography

\usepackage{graphicx}
\usepackage{subfigure}
\usepackage{caption}

\usepackage{wrapfig}

\usepackage{mathtools}
\usepackage{lineno}
\usepackage{color}
\usepackage{comment}
\usepackage{algorithm}
\usepackage{algorithmic}
\usepackage{amsthm}

\usepackage{flushend}
\DeclareMathOperator*{\argmin}{argmin}
\DeclareMathOperator*{\argmax}{argmax}

\usepackage{balance}
\usepackage{enumitem}

\usepackage{capt-of}

\newcommand{\eat}[1]{}
\newcommand{\techrep}[1]{}

\newtheorem{theorem}{Theorem} 
\newtheorem{lemma}[theorem]{Lemma}
\newtheorem{observation}{Observation}

\newtheorem{example}{Example}

\usepackage{array}
\usepackage{xspace}
\usepackage{makecell}

\newcommand{\stitle}[1]{\noindent{\bf #1}}
\newcommand\abol[1]{\textcolor{orange}{(Abol) #1}}
\newcommand\hadis[1]{\textcolor{orange}{(Hadis) #1}}

\newcommand{\new}[1]{#1}

\newcommand\attrib[1]{{\tt \small #1}\xspace}

%% file: intro.tex
\vspace{-3mm}
\section{Introduction}\label{sec:intro}
%\vspace{-3mm}

%-- why responsible data science? --

Data-driven decision making plays a significant role in modern societies
%Data science and advanced computational methods provide an opportunity to
by enabling wise decisions and to make societies more just, prosperous, inclusive, and safe.
However, this comes with a great deal of responsibilities as improper development of data science technologies can not only fail but make matters worse.
% For example, judges in the US courts use background information of the individuals for setting bails or sentencing criminals.
Judges in US courts, for example, use criminal assessment algorithms that are based on the  background information of individuals for setting bails or sentencing criminals.
% This is useful as it can lead to safer societies, but at the same time has the potential to significantly impact the lives of individuals if not developed properly.
While it could potentially lead to safer societies, an improper usage
could result in deleterious consequences on people's lives.
For instance, the recidivism scores provided for the judges are highly criticized as being discriminatory, as they assign higher risks to African American individuals~\cite{propublica}.

Machine learning (ML) is in the center of data-driven decision making as it provides insightful unseen information about phenomena based on available observations.
% Blindly applying machine learning without paying attention to societal impacts, however, can lead to disruptive biased decision making, propagating racism or sexism.
Two major reasons of unfair outcome of ML models are {\em Bias in training data} and {\em Proxy attributes}. The former is mainly due to the inherent bias (discrimination) in the historical data that reflects unfairness in society. For example, redlining is a systematic denial of services used in the past against specific racial communities, affecting historical data records~\cite{redlining}. Proxy attributes on the other hand, are often used due to the limited access to labeled data, especially when it comes to societal applications. For example, when actual future recidivism records of individuals are not available, one may resort to information such as ``prior arrests'' that are easy to collect and use it as a proxy for the true labels, albeit a discriminatory one.

\eat{\techrep{
The following are two major reasons for this to happen:
% That mainly happens due to the two following reasons:
\begin{itemize}[leftmargin=*]
    \item {\em Bias in training data:} ML models use the background information of individuals, which is typically biased due to historical discriminations. For example, redlining is a systematic denial of services used in the past against (mainly) specific racial communities,
    leaving its footprint up to the day and the existing data records~\cite{redlining}.
    % Though redlining has to residents of specific communities with specific racial demographics, often racially associated, neighborhoods or communities, either directly or through the selective raising of prices., leaving their footprint up to the day~\cite{redlining}.
    Gender bias in Data~\cite{datasexism}, including health care~\cite{GBinHealthAI} is yet another example of bias in training data.
    \item {\em Proxy attributes:} labeled data is the corner-stone of supervised learning. Yet, in societal applications there is often  limited labeled data. For example, the aforementioned recidivism scores are meant to show how likely an individual is to commit a crime in the future.
    Similarly, in the context of college admission the goal is to admit students who are likely to be the most successful in the future.
    Training data with such information is either not available or a very limited number of labeled data exists. As a consequence, other available attributes are typically used as proxies for the true labels. For instance, ``getting arrested by cops'' may be considered as a proxy for committing a crime, or GPA for future college success.

\end{itemize}}
\techrep{ML models rely on the data that can be highly biased.
Bias in data can cause model unfairness which can give rise to discrimination in consequent decisions.
% The decisions made from the output of a model based on a biased labeled dataset can result in discrimination and unfairness.
For instance, a job platform can rank less qualified male candidates higher than more qualified female candidates \cite{lahoti2019ifair}.}}

\begin{example}\label{ex1}
% To help judges make a wise decision when setting bails, a company is interested to create a model to predict recidivism; that is, how likely an individual is to commit a crime in the future.
A company is interested in creating a model for predicting recidivism
to be used by judges when setting bails; they want to predict how likely a person is to commit a crime in the future.
Suppose the company has access to the background information of some criminal defendants\footnote{In the US, such information is provided by sheriff offices of the counties. For instance, for the COMPAS dataset, ProPublica used information obtained from the Sheriff Office of the Broward County. \url{https://bit.ly/36CTc2F}}.
However, the collected data is not labeled. %, besides being limited in the count.
% That is because the information about whether or not an individual will commit a crime in the future is not available at the time of trial.
That is because there is no evidence available at the time of the trial if an individual will commit a crime in the future or not.
Considering a time window, it is possible to label an individual in the dataset by checking the background of the individual within the time window after being released. However, it is costly as it might require expert efforts for data collection, integration and entity resolution.
\end{example}

% \techrep{
\begin{example}\label{ex2}
A loan consulting company would like to create a model for financial agencies to identify ``valuable customers'' who will pay off their loans on time.
The company has collected a dataset of customers who have received a loan in the past few years.
The dataset includes information such as demographics, education and income level of individuals.
Unfortunately, at the time of approving loans, it is not known whether customers will pay their debt on time, and hence, the data are not labeled.
% the dataset is not labeled. That is because at the time of approving loans it is not known if the customer will pay his/her debt on time.
% Still, the company has hired experts that given the information of an individual who has received a loan in the past,can check his/her background and identify if the payments have been made on time.
Nevertheless, the company has hired experts who, given the information of admitted applicants in the past, can verify their background and assess if payments were made on time.
Of course, considering the costs associated with a background check, it is not viable to freely label all customers.
\end{example}
% }

Both of the above examples
%}
%\submit{
%Applications similar to Example~\ref{ex1}
%}
use biased historical data for building their models.
For instance, the \attrib{income} in Example~\ref{ex2} is known to include gender bias \cite{jones1983sources}. Similarly, using \attrib{prior count} as a proxy attribute in Example~\ref{ex1} is racially biased \cite{propublica}.
Also, in both examples the datasets are unlabeled.

A new paradigm of \emph{fairness} in machine learning~\cite{fairmlbook} has emerged to address the unfairness issues of predictive outcomes.
These work often assume the availability of (possibly biased) \emph{labeled data} in sufficient quantity.
When this assumption is violated, their performance degrades.
In many practical societal applications such as Example~\ref{ex1},
one operates in a constrained environment.
Obtaining accurate labeled data is expensive, and could only be obtained in a limited amount.
Training the model by using the (problematic) proxy attribute as the true label\footnote{In the rest of the paper we refer to true label as label.}
will result in an unfair model.

In this work, our goal is to develop efficient and effective algorithms for
fair models in an environment where the budget for obtained labeled data is bounded.
%An obvious baseline is to randomly select a subset of data (depending on the available budget),
%obtain the labeled data and use it for training.
% This model would be fairer than the one using proxy attributes.
%However, a more sophisticated approach would be to use an adaptive sampling strategy.
% -- briefly explain different AL strategies\\
% Providing label for many sophisticated supervised learning tasks is difficult and expensive such as...
% In such cases, allowing the model to choose from its unlabeled training set will minimize the labeling cost.
Active learning~\cite{settles2009active} is a widely used strategy for such a scenario.
It sequentially chooses the unlabeled instances where their labeling is the most beneficial for the performance improvement of the ML model.
% By default, active learning selects tuples that improves
% the predictive accuracy of the model and hence could sustain or intensify unfairness of the model.
% Despite its importance, to the best of our knowledge, none of the existing work in active learning takes {\em fairness} into account.
% That is indeed our purpose in this paper.

In this paper, we aim to develop an active learning framework that will yield fair(er) models.
Fairness has different definitions and is measurable in various ways.
Specifically, we consider a model fair if its outcome does not depend
on sensitive attributes such as race or gender.
We adopt demographic parity (aka statistical parity), one of the popular fairness measures \cite{kusner2017counterfactual,dwork2012fairness}.
\eat{
Although we consider model independence as our measure of fairness, in \S~\ref{sec:disc} we demonstrate how to extend our findings for other notions of fairness based on separation and sufficiency~\cite{fairmlbook}.
}

\stitle{Summary of contributions.}
We introduce {\em fairness in active learning} for constructing fair models in the context of limited labeled data.
\new{
We propose a fair active learning (FAL) framework to balance model performance and fairness (\S~\ref{sec:framework}). 
At a high-level, FAL uses an accuracy-fairness optimizer for selecting samples to be labeled. 
We propose three strategies for the optimizer, namely {\bf FAL $\alpha$-aggregate} (\S~\ref{sec:opt:alpha}), {\bf FAL Nested} (\S~\ref{sec:opt:nested}), and {\bf FAL Nested Append} (\S~\ref{sec:opt:nAppend}).
Given that sample points are unlabeled in the context of AL,
the optimizer uses {\em expected unfairness reduction}, proposed in \S~\ref{sec:framework}.
For the special case of generalized linear models, we propose
an {\em fairness by covariance} (\S~\ref{sec:fbc}), an efficient alternative for expected unfairness reduction that reduces the asymptotic time complexity of FAL to be the {\em same} as traditional active learning.
While our default notion of fairness is based on Demographic Parity, we also provide extension to other fairness models (\S~\ref{sec:disc}).
We conduct comprehensive experiments to
evaluate the performance of our proposal on benchmark datasets (\S~\ref{sec:exp}).
Our result confirm that FAL can significantly reduce unfairness while not significantly impacting the model accuracy.
In particular, our optimization {\bf FAL Nested Append} had the best performance across different experiments and fairness models, confirming the improvements proposed for accuracy-fairness trade-off.
}%\abol{add a couple of sentences to highlight takeaway from exp.}
% show that performing active learning while considering the fairness constraint can significantly improve the fairness of a classifier without any major reduction in model accuracy.
% \hadis{this will be revised based on new results}
%Our experiment results across different fairness metrics confirm significant improvements in fairness without major reductions in model accuracy.
\eat{
In summary, our contributions are as following:
\begin{itemize}
    \item
    Carefully formalizing terms and background, we present different fairness measures based on model independence.
    \item
    We introduce fairness in active learning, an iterative approach that incorporates the fairness measure in its sample selection unit and constructs a fair predictive model as a result.
    % \item We show that changing the training data changes demographic parity.
    \item We propose the expected fairness measure for unlabeled sample points based on the best-known estimate of the function.
    \item
    We conduct comprehensive experiments on real-world data, considering different fairness metrics based on model independence. Our results show an improvement of around 50\% in fairness measures while not significantly impacting the model accuracy.
    \item We discuss how to extend our framework for different fairness measures.
\end{itemize}
In the remainder of this paper, we start with a background on active learning and our fairness model in \S~\ref{sec:back}. In \S~\ref{sec:alg}, we will present the FAL framework.
In \S~\ref{sec:disc}, we will discuss how to extend our proposal for different fairness measures.
We will provide our comprehensive experiments in\S~\ref{sec:exp},
review the related work in \S~\ref{sec:related},
and conclude the paper in \S~\ref{sec:con}.
}

%% file: related_work.tex
\vspace{-3mm}
\section{Related work}\label{sec:related}

% To the best of our knowledge, none of the existing active learning works considers fairness of the predictive outcome.
% There is a comprehensive survey~\cite{settles2009active} on different active learning scenarios (Membership Query Synthesis, Stream-Based Selective Sampling, pool-Based Active Learning) and sampling strategies (Uncertainty Sampling, Query-By-Committee, Expected Model Change, Variance Reduction, etc.).
% \hadis{probably better to remove application papers and put more theoretical papers like the ones reviewers recommended}. \textcolor{blue}{Sara: agreed. will find some such works.} Active Learning has been widely used in different applications for training a wide range of classifires, where the labeling process for these datasets is labor-intensive and costly. The examples are
% image and speech recognition~\cite{hoi2006batch,joshi2009multi,minakawa2013image,yu2010active,riccardi2005active},
% information retrieval~\cite{tian2011active},
% text analysis\cite{tong2001support,cormack2016scalability,hu2016active,davy2007dimensionality}, recommender systems \cite{sun2013learning,resnick1997recommender,houlsby2014cold}.

\stitle{Algorithmic Fairness in ML.}
Algorithmic fairness is a topic of extensive interest with
% There is a body of work in designing fair algorithms using different fairness definitions.
\cite{barocas2017fairness,vzliobaite2017measuring,mehrabi2019survey} providing surveys on discrimination and fairness in machine learning.
Existing works have formulated fairness in classification as a constrained optimization~\cite{zafar2017fairness,menon2018cost,celis2019classification,hardt2016equality,huang2019stable}.
% Fairness in classification has been formulated as a a constrained optimization problem, to incorporate fairness constraints, \cite{zafar2017fairness,zafar2015fairness,menon2018cost,corbett2017algorithmic,celis2019classification,hardt2016equality,huang2019stable,asudeh2019designing}.
A body of work focus on modifying the classifier, in-process, to build fair classifiers~\cite{fish2016confidence,goh2016satisfying,dwork2012fairness,komiyama2018nonconvex,corbett2017algorithmic}.
Some others remove disparate impact through pre-processing the training data~\cite{zemel2013learning,kamiran2012data,feldman2015certifying,krasanakis2018adaptive,asudeh2019assessing}, while the last group post-process model outcomes to achieve fairness \cite{kim2018multiaccuracy,hardt2016equality,pleiss2017fairness,hebert2017calibration}.
{\em Our proposal is orthogonal to the fair ML literature.}
While our goal in this paper is on selecting the samples to be labeled,
fair ML algorithms aim to build fair models for a
given set of labeled samples.
% As we shall show in \S~\ref{sec:exp}, the fair ML algorithms can be integrated with our proposal for building the model.

%Sara: made a pass and kept the original in case you want to revert.
%\hadis{To the best of our knowledge, there is only one work that considers fairness in active learning \cite{sharafpromoting}. Although, they used our original FAL approach as a baseline, they had a different setting with the assumption of available large enough set of labeled pool for validation. Fairness has also been studied in few works that consider the intersection of fairness and active feature acquisition \cite{noriega2019active,bakker2020fair}. }
\new{ 
This paper is the first to introduce fair active learning.
Besides this paper, we are aware of one subsequent work that considers fairness in active learning \cite{sharafpromoting}. The setting of this work differs from the standard active learning setting: instead of seeking to minimize the number of labeled data, \cite{sharafpromoting} starts with a pre-existing labeled data and seek to minimize the disparity by labeling \emph{additional} samples. 
In contrast, we tackle the traditional active learning setting with no labeled data.
Fairness has also been studied in few works that consider the intersection of fairness and active feature acquisition \cite{noriega2019active,bakker2020fair}. 
Our work is orthogonal to this research since our goal is not feature acquisition but rather active learning.
}
% \abol{I made some changes in the prev. paragraph. Please take a careful look.}

There has been extensive work in the KDD community on diverse aspects of fairness including
quantification and evaluation~\cite{speicher2018unified,srivastava2019mathematical,diciccio2020evaluating},
tradeoffs of fairness~\cite{corbett2017algorithmic} and
applications such as ranking~\cite{beutel2019fairness},
search and recommenders~\cite{geyik2019fairness} and
truth discovery~\cite{li2020towards}.

%S: commented this out and replaced it with KDD specific references
%Fairness in data-driven decision making has also been studied in related topics, such as ranking~\cite{asudeh2019designing,guan2019mithraranking,yang2018nutritional,zehlike2017fa} and  recommendation systems~\cite{burke2017multisided,tsintzou2018bias,yao2017beyond}.

%S: moved to the bottom to emphasize the fairness first.
\stitle{Active Learning.}
Different active learning scenarios (Membership Query Synthesis, Stream-Based Selective Sampling, pool-Based Active Learning) and sampling strategies (Uncertainty Sampling, Query-By-Committee, Expected Model Change, Variance Reduction, etc.) have been proposed and are surveyed in \cite{settles2009active}. Uncertainty sampling is one of the most popular approaches for active learning~\cite{lewis1994heterogeneous,balcan2007margin,tong2001support}, which merely selects data points based on the single objective function of informativeness. There are several active learning approaches proposed to incorporate more than one criteria for sampling, such as representativeness~\cite{xu2003representative,donmez2007dual,huang2010active}.

% Fairness has also been studied in
% special ML context such as reinforcement learning~\cite{jabbari2017fairness}, adversarial networks~\cite{wadsworth2018achieving,xu2018fairgan}, and feature acquisition~\cite{noriega2019active}.
% Fairness in data-driven decision making has also been studied in related topics, such as ranking~\cite{asudeh2019designing,guan2019mithraranking,yang2018nutritional,zehlike2017fa}
% and  recommendation systems~\cite{burke2017multisided,tsintzou2018bias,yao2017beyond}.

% Dwork et.al~\cite{dwork2012fairness} addressed fairness in classification maximizing utility subject to the fairness constraint, focusing on equal treatment of individuals from different groups.

% Huang et.al~\ref{huang2019stable} proposed a stability-focused regularization term in the fair classification optimization formulation to construct a fair classifier that is stable with respect to variations in the
% training dataset.

%% file: background.tex
\section{Background}\label{sec:back}
In this section, we introduce the data model, the active learning framework with
uncertainty sampling heuristic, and fairness model.

\subsection{Learning Model}\label{sec:back-AL}

Given a classifier and a pool of unlabeled data $\mathcal{U}$,
Active Learning (AL)  identifies the data points to be labeled so that
an accurate model could be learned as quickly as possible.
% \hadis{I put this back based on reviewer comment}
$\mathcal{U}$ is assumed to be an independent and identically distributed (i.i.d) sample set collected from the  underlying unknown distribution.
For each data point $P_i\in\mathcal{U}$,
we use the notation $X^{(i)}$ for the $d$-dimensional vector of input features
and $X^{(i)}_j$ to refer to the value of $j$th feature, $x_j$.
Each data point is associated with a non-ordinal categorical sensitive attribute $S$
such as \attrib{gender} and \attrib{race}.
We use the notation $S^{(i)}$ to refer to the sensitive attribute of $P_i$.
We also use $y^{(i)}$ to refer to the label of a point $P_i$ with $K$ possible values
%of the label space $Y=\{0, \cdots, K-1\}$.
$\{0, \cdots, K-1\}$.
%We assume an unknown underlying distribution over input space $X$, consisting of $d$ features $\{x_1, x_2, \cdots,x_d\}$, from which we collected an independent and identically distributed (i.i.d.) \emph{unlabeled pool} of samples $\mathcal{U}$.
%During the training process, the subset $\mathcal{L}\subseteq \mathcal{U}$ consists of data points for which labels are provided by the labeling oracle is referred to as {\em labeled pool}.
%We use the pair $\langle X^{(i)},S^{(i)}\rangle$ for a point $P_i$ in $\mathcal{U}$, and the triple $\langle X^{(i)},S^{(i)},y^{(i)}\rangle$ for every point in the labeled pool $\mathcal{L}$.

\eat{
% \begin{wrapfigure}{R}{0.5\textwidth}
% %\vspace{-8mm}
%     \begin{minipage}{0.5\textwidth}
      \begin{algorithm}[H]
        \caption{{\bf Active Learning \textcolor{red}{S: Is this necessary?}\abol{no}}}
        \begin{algorithmic}[1]
        \label{alg:al}
        \FOR{$t=1$ to $B$}
            \STATE $X^* = \underset{X\in \mathcal{U}}{\mbox{argmax}}~ \mathcal{H}(y|X,\mathcal{L})$
            \STATE $y =$ label $X^*$ using the labeling oracle
            \STATE add $\langle X^*,y\rangle$ to $\mathcal{L}$
            \STATE train the classifier $C_t$ using $\mathcal{L}$
        \ENDFOR
        \STATE {\bf return} $C_B$
        \end{algorithmic}
      \end{algorithm}
%     \end{minipage}
%     %\vspace{-3mm}
%   \end{wrapfigure}
}

% Without loss of generality, in this paper we assume $S$ is a single binary attribute. In \S~\ref{sec:discussions}, we will discuss how to generalize for multiple non-binary sensitive attributes.
%We use the notation $S^{(i)}$ to refer to the sensitive attribute of $P_i$.
%\eat{Without loss of generality and to simplify the explanations, unless explicitly stated, we assume $S$ is a single sensitive attribute. Still, we would like to emphasize that our techniques are not limited to the number of sensitive attributes.}
%\abol{if we have space, we can discuss the (easy) extension to multiple sensitive attributes in the discussions section}
%We also use $y^{(i)}$ to refer to the label of a point $P_i$ with $K$ possible values of the label space $Y=\{0, \cdots, K-1\}$.
%\eat{
%We assume the labels of the data points are initially unknown.
%In \S~\ref{sec:back-AL}, we will explain how to obtain the label of a data point $P_i$.

%At any moment during the training process, the subset $\mathcal{L}\subseteq \mathcal{D}$ of data points for which labels are known is referred to as {\em labeled pool} and the rest of them $\mathcal{U}=\mathcal{D}-\mathcal{L}$ is called {\em unlabeled pool}.
%}
%Finally, we use the pair $\langle X^{(i)},S^{(i)}\rangle$ for a point $P_i$ in $\mathcal{U}$, and the triple $\langle X^{(i)},S^{(i)},y^{(i)}\rangle$ for every point in the labeled pool $\mathcal{L}$.

% For many real-world problem the learner has access to a pool of unlabeled data points.

The goal is to learn a classifier function $C:X\rightarrow Y$
that maps the feature space $X$ to the labels $Y$.
Let $\hat{y} = C(X)$ be the predicted label for $X$.
%
% In other words, given a data point with features $X$, $\hat{y} = C(X)$ is the predicted label for $X$ based on $C$.
% All the data points are initially unlabeled.
Pool-based active learning~\cite{lewis1994sequential}, sequentially selects instances from  $\mathcal{U}$ to be labeled by an {\em expert oracle} and forms a labeled set $\mathcal{L}$ for training.
%Active learning assumes the existence of an {\em expert oracle} that given a data point $P$ provides its label.
% \abol{provide some examples here}
Labeling, however, is costly and usually there is a {\em limited labeling budget} $B$.
% In active learning, we have a limited collection of labeled set of examples $L = \{(x_1,y_1),\dots,(x_m,y_m)\}$ where for each example $x_i \in \mathbb{R}^d$ represents a set of features and the scalar $y_i$, is a label indicating whether an example belongs to a class ranges over all possible labeling, $\{1,\dots,K\}$.
% With such data set we can build a classifier $C(x)$, however, the goal is to fit a classifier that performs well with the minimum number of labeled data.
%Using the sampling budget, one can randomly label $B$ data points and utilize them to train a classifier.
The challenge is to design an effective sampling strategy that wisely utilizes the budget to build the most accurate model.
% Hence, the challenge is to find a smart sampling strategy to choose good candidates from the pool.
%In pool based sampling all of the unlabeled data points will be evaluated at once.
%
%Different sampling strategies have been proposed in the context of active learning.
Uncertainty sampling~\cite{lewis1994sequential}, a widely used strategy, % in active learning for classification.
% For classification the choice of the next sample points with uncertainty sampling can be viewed as the instances about which we are least certain how to label.
%It selects data points for labeling such that the model variance is maximally minimized.
chooses the point $P\in\mathcal{U}$ that the current model is least certain about its label.
The classifier $C_{t-1}$ for iteration $t$ chooses the data point that maximizes
the Shannon entropy ($\mathcal{H}$)~\cite{shannon1948mathematical} over the label probabilities.
%At every iteration of the process, let the classifier $C$ be the current model, using the labeled dataset $\mathcal{L}$.
%For every data point $P_i\in\mathcal{U}$ with the feature vector $X^{(i)}$, let $\mathbb{P}(y=k|X^{(i)})$ be the posterior probability that its unknown label $y^{(i)}$ will be $k$ based on $C$.
%
%By maximizing the uncertainty, active learning selects the points that are close to the decision boundary, where we are least certain about the class label. Uncertainty can be defined in different ways~\cite{settles2009active}. In general, uncertainty sampling refers to maximum entropy~\cite{shannon1948mathematical}. Equation \ref{eq:entropy} denotes the Shannon entropy formulation for classifying $X\in U$ with $y\in\{0,1\}$ based on the probability obtained from the classifier $C$ for an unknown label variable $y$.
\begin{equation}\label{eq:entropy}
    X^* = \underset{X\in \mathcal{U}}{\mbox{argmax}}~ \mathcal{H}(y|X,\mathcal{L})
\end{equation}
%\vspace{-3mm}
%where $\mathcal{H}$ is defined as follows:
%\begin{equation}
%    \mathcal{H}(y|X,\mathcal{L})= - \sum_{k=0}^{K-1} \mathbb{P}(y=k|X, \mathcal{L})\log \mathbb{P}(y=k|X, \mathcal{L})
%\end{equation}

% \techrep{
% Algorithm~\ref{alg:al} presents the standard active learning algorithm, using Equation~\ref{eq:entropy}.
% Iteratively, the algorithm selects a point from $\mathcal{U}$ to be labeled next.
% }
% \submit{
%\vspace{-2mm}
Using Equation~\ref{eq:entropy}, the active learning algorithm
iteratively selects a point from $\mathcal{U}$ to be labeled next.
% }
It uses the classifier trained in the previous step to obtain probabilities of the labels.
%$\mathbb{P}(.)$ (initially, all the probabilities are equal), and to calculate the entropies.
The algorithm obtains the label from the labeling oracle, and adds the point to the set of labeled dataset $\mathcal{L}$, using it to train the classifier $C_t$.
%It uses $\mathcal{L}$ to train the classifier $C_t$, where $t$ shows the current iteration.
This process continues until the labeling budget is exhausted.

\eat{
\begin{table*}[!tb]
    \centering
    \begin{tabular}{|c||c|c|}
    \hline
         & $\mathcal{F(S,C)}_I=I(\hat{y};S)$ & $|cov(\hat{y},S)|$  \\ \hline
         difference& $|\mathbb{P}(\hat{y}=1|S=0) - \mathbb{P}(\hat{y}=1|S=1)|$ &  $|\mathbb{P}(S=1|\hat{y}=1) - \mathbb{P}(S=1)|$\\ \hline
         ratio & $1 - \min\Big(\frac{\mathbb{P}(\hat{y}=1|S=0)}{\mathbb{P}(\hat{y}=1|S=1)},\frac{\mathbb{P}(\hat{y}=1|S=1)}{\mathbb{P}(\hat{y}=1|S=0)}\Big)$
         &  $1-\min\Big(\frac{\mathbb{P}(S=1|\hat{y}=1)}{\mathbb{P}(S=1)},\frac{\mathbb{P}(S=1)}{\mathbb{P}(S=1|\hat{y}=1)}\Big)$\\ \hline
    \end{tabular}
    \caption{Difference metrics for measuring demographic disparity.}
    \label{tab:DemoMeasures}
\end{table*}
% \footnotetext{In addition to the measures in Table~\ref{tab:DemoMeasures}, one could use correlation $|corr(\hat{y},S)|$, or as in~\cite{agarwal2018reductions} $|\mathbb{P}(\hat{y}=1|S=1) - \mathbb{P}(\hat{y}=1)|$ for measuring disparity.}
} % end of techrep
%\vspace{-3mm}
\subsection{(Un)Fairness Model}\label{sec:back-FM}
%\vspace{-2mm}
We develop our fairness model on the notion of {\em model independence} or {\em demographic parity} (DP) \cite{fairmlbook,vzliobaite2017measuring,zafar2017fairness}, also referred by terms such as statistical parity \cite{dwork2012fairness,simoiu2017problem}, and disparate impact \cite{barocas2016big,feldman2015certifying}.
\new{
Although our focus in this paper is on fairness based on model independence, in \S~\ref{sec:disc} we show how to extend our framework for other measures based on separation ($\hat{y}\bot S~|~ y$) and sufficiency ($y\bot S~|~ \hat{y}$) \cite{fairmlbook}.
}
%
%Apparently, disparities in the model does not necessarily imply the designer's wish to be unfair. The problem occurs as these models rely entirely on (biased) historical data for learning a system; therefore, the historical disparities in the data cause the (unintentional) bias in the model.
%Sara: removed this non-sequitur
%We believe that machine learning practitioners are responsible to intervene in the modeling process (in different learning stages) to mitigate model disparities.
%
Given a
classifier $C$ and a
random point $\langle X,S\rangle$ with a predicted label $\hat{y} = C(X)$, DP holds iff $\hat{y}\bot S$ \cite{barocas2017fairness,fairmlbook}.
For a binary classifier, let $\hat{y}=1$ count as ``acceptance'' (such as receiving a loan).
%\techrep{ -- in Example \ref{ex2}, the group that receive a loan}.
DP requires that the acceptance rate be the same for all groups of $S$ i.e. female or male in this case. For a binary classifier and a binary sensitive attribute, %all of the followings hold:
the statistical independence of a sensitive attribute from the predicted label induces the following notions for DP:

\begin{enumerate}[leftmargin=*]
    \item $\mathbb{P}(\hat{y}=1|S=0) = \mathbb{P}(\hat{y}=1|S=1)$:
    The probability of acceptance is equal for members of different demographic groups.
    For instance, in Example~\ref{ex1} members of different race groups have an equal chance for being classified as low risk.
    \item $\mathbb{P}(S=1|\hat{y}=1) = \mathbb{P}(S=1)$:
    If the population ratio of a particular group is $\rho$ (i.e. $\mathbb{P}(S=1))$, the ratio of this group in the accepted class is also $\rho$.
    For instance, in Example~\ref{ex2}, let $\rho$ be the female ratio in the applicants' pool. Under DP, female ratio in the set of admitted applications for a loan equals to $\rho$.
    \item $I(\hat{y};S) = 0$:
    Mutual information is the measure of mutual dependence between two variables.
    When $\hat{y}$ and $S$ are independent, their mutual information is zero. That is, the conditional entropy $H(S|\hat{y})$ is equal to $H(S)$.
    \item $cov(\hat{y},S) = 0$:
    {When $\hat{y}$ and $S$ are independent, $cov(\hat{y},S)$ is equal to zero.}
    \eat{The covariance between the target and sensitive features is zero. Under DP $\hat{y}$ and $S$ are independent. As a result, the covariance $cov(\hat{y},S)$ is equal to zero.}
\end{enumerate}

A disparity (or unfairness) {\em measure} can be defined using any of the above notions.
\eat{
Table~\ref{tab:DemoMeasures} \hadis{we need to name the measure and use F notation} summarizes some of the ways the disparity can be measured. As stated in the first row of the table, mutual information and covariance (or correlation) provide two natural measures.
Another way of quantifying the disparity is by subtracting the probabilities (Row 2 of Table~\ref{tab:DemoMeasures})
or the ratios between probabilities (Row 3 of Table~\ref{tab:DemoMeasures}).
Consistent with Row 2, we defined the ratio-based measures such that zero is the maximum fairness and the measure is in the range [0,1].
}
{
The absolute differences or the ratio of probabilities in bullets 1 or 2 provide four disparity measurements.
Mutual information and covariance (or correlation) provide two natural measures, since the greater the absolute value of the two is the greater the disparity.
}
\eat{
Finally,
we would like to reiterate that we consider the societal norms, which are not always quite aligned with statistical measures.
Due to the societal discrimination against some minority groups, social data is usually ``biased'' \cite{olteanu2019social}.
Hence, actions known as reverse discrimination (such as affirmative action) are taken to increase the presence of underrepresented groups in the outcomes.
Such fairness guidance is usually provided by law.
That can be viewed as a higher acceptance probability for certain protected groups of sensitive attributes.
}
In this paper, we do not limit ourselves to any of the unfairness measures (demographic disparity) and give the user the freedom to provide a customized measure.
%
%
%In other words, {\em we are agnostic to the choice of fairness measure}.
% It is worth mentioning that although we used a single binary sensitive attribute for the explanations and measures, a measure can be defined over multiple non-binary sensitive attributes with overlapping protected groups.
%Generalizing our notion of unfairness to any measure based on demographic disparity, in the rest of the paper,
We denote (user-provided) measure of unfairness as $\mathcal{F}(S,C)$.
When $S$ is clear by context, we simplify it to $\mathcal{F}(C)$.

% \abol{add a section on how to balance all measures by satisfying the ...}

% \abol{can we still have the population bias component?
% Can we use techniques from label balancing to formalize population ratio balancing?
% }

% \techrep{
% It is worth mentioning that, algorithmic fairness can be achieved by intervention at pre-processing, in-processing, or post-processing strategies \cite{friedler2019comparative}. This paper provides an {\em in-process} strategy for fairness.
% }

%% file: tech/framework.tex
\section{Fair Active Learning (FAL) Framework}\label{sec:framework}

% \abol{the following needs a heavy work to connect this and next section}

By carefully selecting samples to be labeled,
AL has the potential to mitigate algorithmic bias by incorporating the fairness measure into its sampling process.
Still, not considering fairness while building models can result in model unfairness.
As a na\"ive resolution, one could decide to drop the sensitive attribute from the training data.
This, however, is not sufficient since the bias in the features may cause model unfairness~\cite{buolamwini2018gender,zou2018ai}. %manrai2016genetic
%As we shall later show in Lemma~\ref{th:cov}, bias in labeled training data $\mathcal{L}$ is the reason for model unfairness.
%The samples selected for labeling in an active learning framework can significantly impact the fairness of a trained model.
Hence, a smart sampling strategy is needed to mitigate the bias.
\eat{
A question which might come to mind is that will any of the existing AL strategies generate fair models. The answer is: \textit{not necessarily}. We show this in Observation~\ref{th:ALnotWork}.
\begin{observation}\label{th:ALnotWork}
An active learning method with any sampling strategy that does not consider fairness may generate unfair models.
\end{observation}
Following Lemma~\ref{th:ALnotWork}, AL strategies that do not consider fairness may generate unfair models.
}
%
%On the other hand, b
Blindly optimizing for fairness could result in an inaccurate model.
%, the model may lose its purpose for accuracy while obtaining high performance is indeed the main objective in any ML framework, including active learning.
For instance, in Example~\ref{ex1}, consider a model that randomly classifies individuals as high-risk. This model indeed satisfies demographic parity since the probability of the outcome is (random and therefore) independent of $S$.
However, such a model provides zero information about how risky an individual is.

\begin{figure}[!t]
\centering
    % \vspace{-5mm}
    \includegraphics[width=0.75\linewidth]{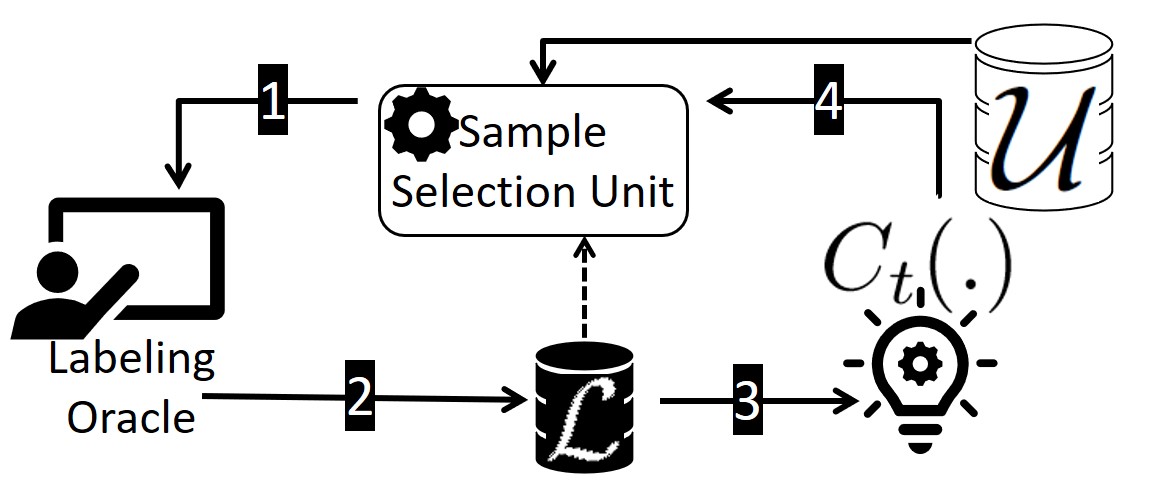}
    \vspace{-5mm}
    \caption{\small FAL framework.}
    \label{fig:arch}
    \vspace{-6mm}
%\end{wrapfigure}
\end{figure}

\new{
We propose the Fair Active Learning (FAL) framework to balance between accuracy and fairness.
}
FAL is an iterative approach similar to standard active learning approaches.
As shown in Figure~\ref{fig:arch}, the central component of FAL is the sample selection unit (SSU) that chooses an unlabeled point $\langle X^{(i)},S^{(i)} \rangle$ from $\mathcal{U}$ and obtains the label from an oracle. The labeled point $\langle X^{(i)},S^{(i)},y^{(i)} \rangle$ is moved to $\mathcal{L}$, the set of labeled points that is used to train $C_t$. %, the classifier at iteration $t$.
In the next iteration, $t+1$, SSU employs $C_t$ and selects the next point to be labeled. This process continues until the budget for labeling is exhausted.

%\begin{wrapfigure}{r}{0.3\textwidth}

\new{
At a high level, SSU can be viewed as two computation blocks stacked on top of each other.
The upper block is the fairness-accuracy optimizer that selects a point from $\mathcal{U}$ to be labeled next such that a combination of accuracy (misclassification reduction) and fairness (unfairness reduction) is maximized.
We shall provide the details of this block in \S~\ref{sec:alg}.
}

\new{
The accuracy-fairness optimizer relies on the lower block for estimating unfairness values.
Let
$C_{t-1}$ be the current model, created in the previous iteration $t-1$.
In order to evaluate the unfairness reduction after labeling a sample point $\langle X^{(i)},S^{(i)} \rangle\in\mathcal{U}$, the optimizer block needs to compute the unfairness of the current model $\mathcal{F}(C_{t-1})$, as well as the unfairness of the model after labeling the sample point $\mathcal{F}(C_{t}^i)$.
Computing $\mathcal{F}(C_{t}^i)$ turns out to be problematic as at the time of evaluating the candidate points in $\mathcal{U}$, we still do not know their labels. On the other hand, to evaluate $\mathcal{F}(C_{t}^i)$, we need to know what the model parameters will be after labeling the point $\langle X^{(i)},S^{(i)} \rangle$ and adding it to $\mathcal{L}$.
In other word, in order to evaluate a point to whether or not it should be labeled, we need to know its label in advance!
This contradicts with the fact that $\mathcal{U}$ is unlabeled.
% This requires knowing {\em all} labels beforehand which contradicts with the fact that $\mathcal{U}$ is unlabeled.
}

% \abol{After thinking a little bit, I came to conclusion that we can keep this fairness measurement method -- our direct computation -- here. Hence the sections after optimization will talk about alternative fairness measurement}

To resolve this issue, using a decision theoretic approach ~\cite{settles2009active}, we consider the \textit{expected unfairness reduction}: selecting the point that is expected to impart the largest reduction to the current model unfairness, \emph{after acquiring its label}. Therefore, instead of $\mathcal{F}(C_{t})$ in Equation~\ref{eq:opt}, we plug in the expected fairness $E\big[\mathcal{F}^{i}_{t}\big]$.  %, we use Equation~\ref{eq:opt2} to pick a sample.
In this way, we are approximating the expected future fairness of a model using $\mathcal{L}\cup X, \forall X\in \mathcal{U}$ \emph{over all possible labels} under the current model.
\eat{In other words, SSU should select the sample point  $X$, which if labeled and added to $\mathcal{L}$ would result in a new model with reduced unfairness.}
Consider a point $P_i\in\mathcal{U}$ and let
$C^{i,k}_t$ be the model after adding $P_i$ to $\mathcal{L}$ if its true label is $y^{(i)}=k$.
Inevitably, SSU does not know the label in advance. Hence, it must instead calculate the unfairness as an expectation over the possible labels.
% \vspace{-2mm}
% \begin{align}
% \label{eq:opt2}
% \argmax_{\langle X^{(i)},S^{(i)}\rangle \in\mathcal{U}} ~~ \alpha\, \mathcal{H}_{t-1}(y^{(i)}) + (1-\alpha)\big(\mathcal{F}(C_{t-1})-E\big[\mathcal{F}^{i}_{t}\big]\big)
% \end{align}
Equation~\ref{eq:E[f]} denotes the expected (un)fairness computation used by SSU:

% \hadis{R: you cannot compute $P(y = k|X)$. Here you should say what you do use, you use the probability according to your model: $P(\hat{y} = k | X_i, X_L, Y_L).$}

%\vspace{0mm}
\begin{align}
\label{eq:E[f]}
E\big[\mathcal{F}^{i}_{t}\big]  = \sum_{k=0}^{K-1} \mathcal{F}(C^{i,K-1}_t) \mathbb{P}(y=k|X^{(i)})
% \vspace{-6mm}
\end{align}

Using Figure~\ref{fig:EofF} for explanation, for every point $P_i=\langle X^{(i)},S^{(i)}\rangle$ in the unlabeled pool,
SSU considers different values of $\{0,\dots,K-1\}$ as possible labels for $P_i$.
For every possible label $y_k$, it updates the model parameters to the intermediate model $C^{i,k}_t$ using $\mathcal{L}\cup \{\langle X^{(i)},  S^{(i)}, k\rangle\}$.

\begin{figure*}[!ht]
    \begin{minipage}[t]{0.38\linewidth}
        	\centering
        	\includegraphics[width =\textwidth]{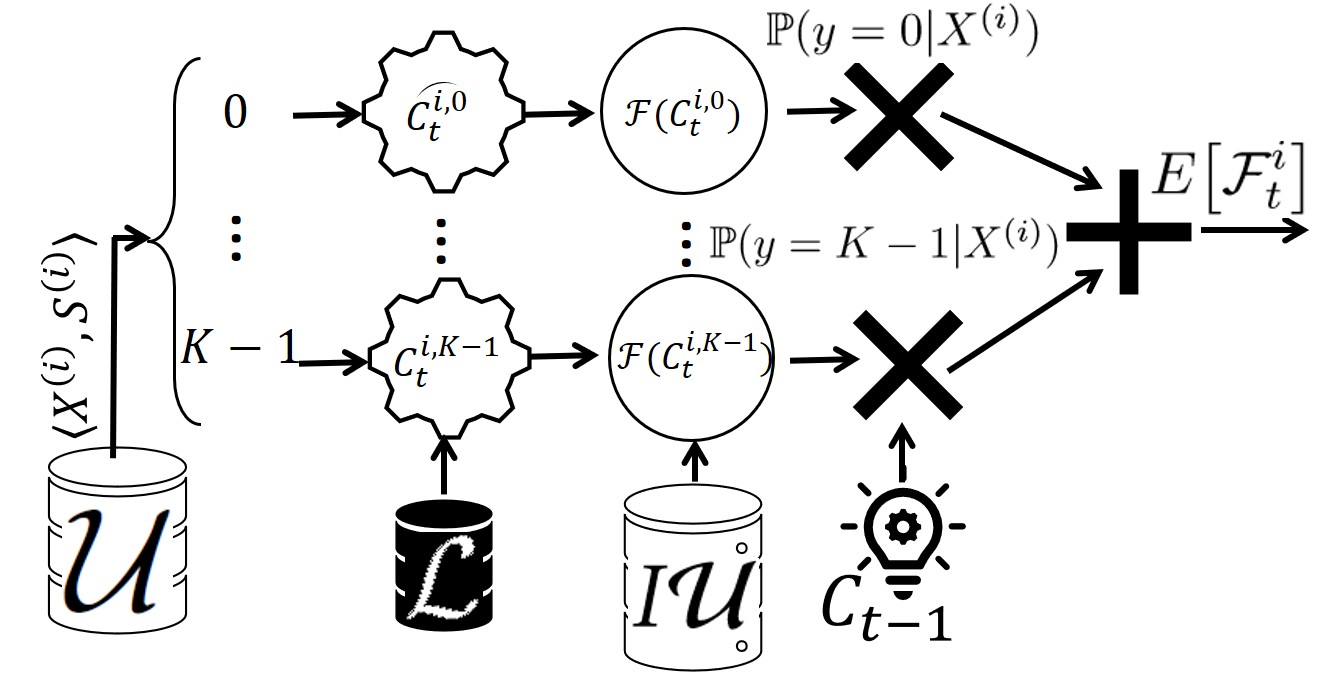}
        	\vspace{-8mm}\caption{\small Computing expected unfairness for $\langle X^{(i)},S^{(i)}\rangle\in\mathcal{U}$.}
    \label{fig:EofF}
    \end{minipage}
    \hfill
    \begin{minipage}[t]{0.33\linewidth}
        	\centering
        	\includegraphics[width =\textwidth]{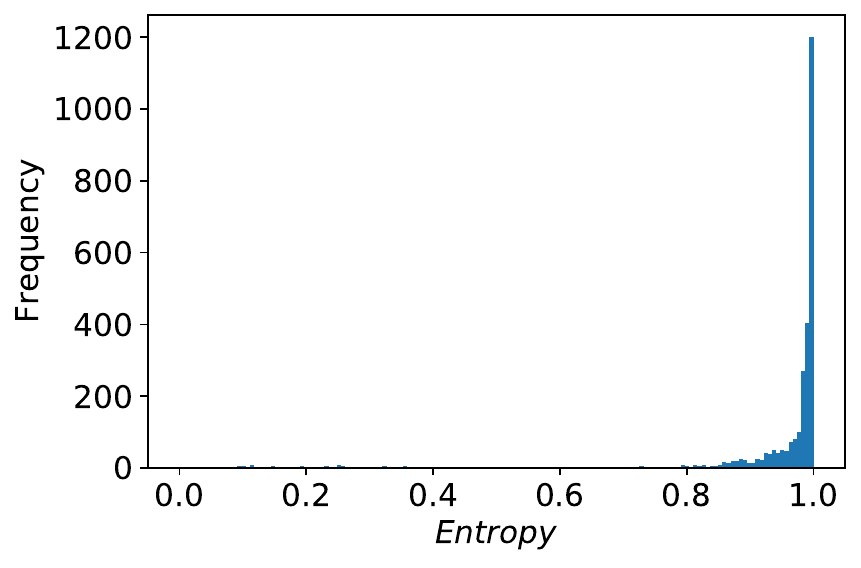}
        	\vspace{-8mm}\caption{\small Distribution of entropy for COMPAS dataset.}
    \label{fig:edist}
    \end{minipage}
    \hfill
    \begin{minipage}[t]{0.25\linewidth}
        \centering
        	\includegraphics[width =\textwidth]{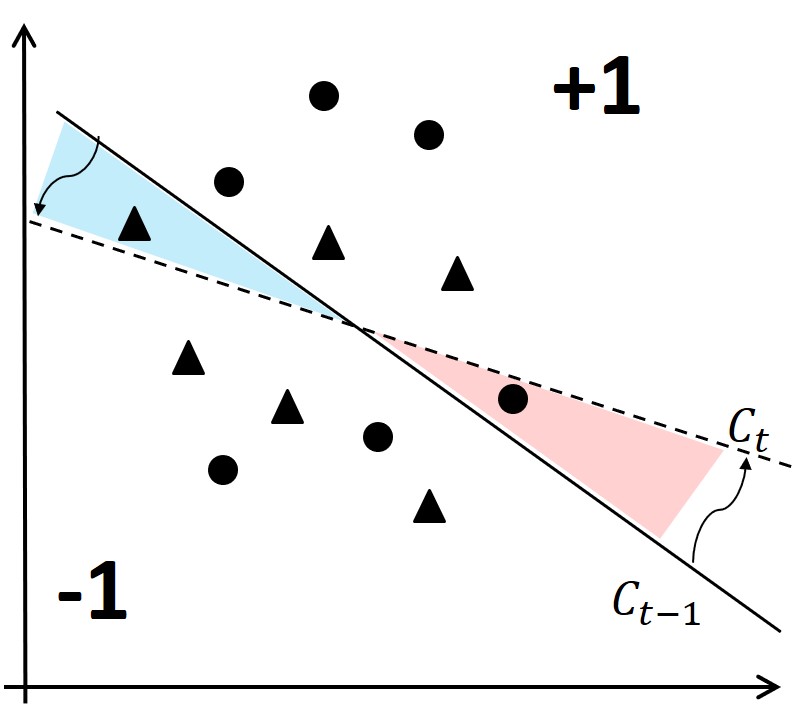}
        	\vspace{-8mm}\caption{\small Toy example.}
    \label{fig:append}
    \end{minipage}
    \vspace{-3mm}
\end{figure*}

\eat{%condense
%\begin{wrapfigure}{r}{0.5\textwidth}
\begin{figure}
\centering
    \includegraphics[width=\linewidth]{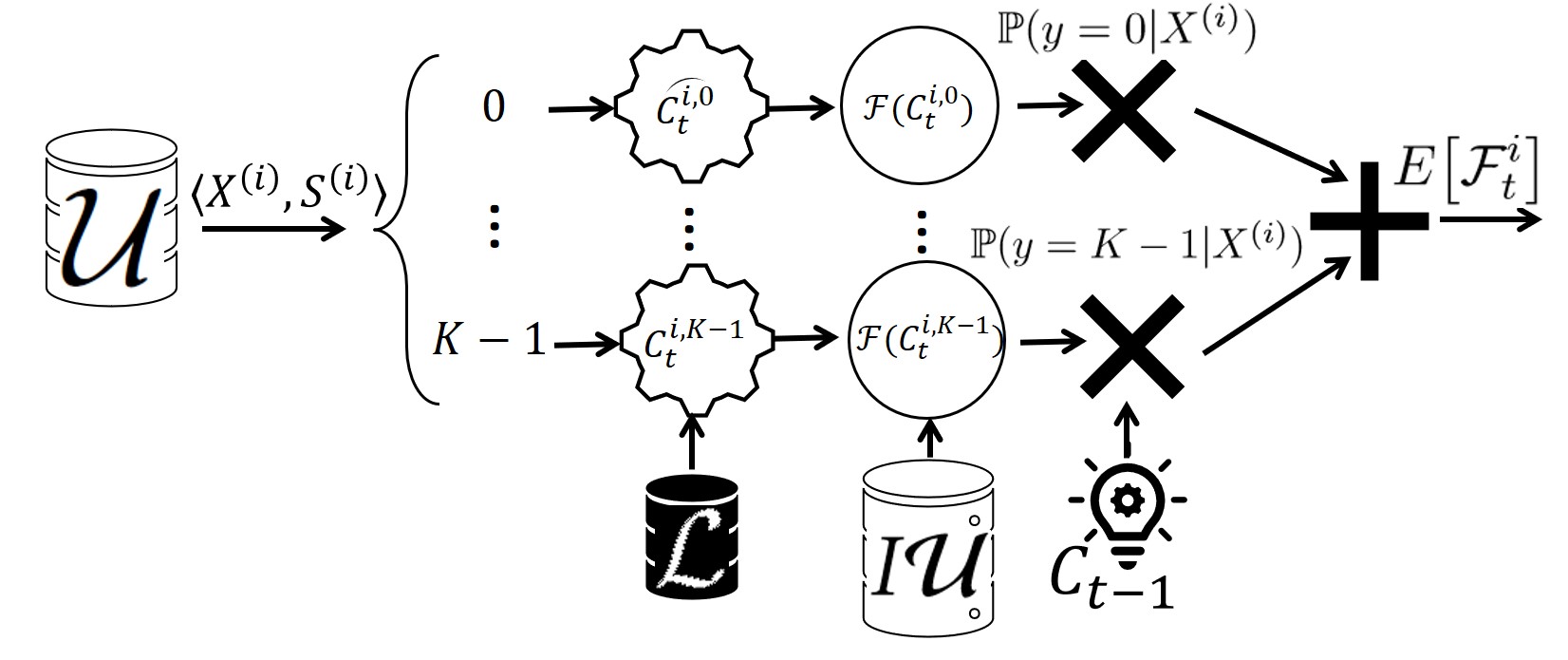}
    \vspace{-8mm}
    \caption{\small Computing expected unfairness for $\langle X^{(i)},S^{(i)}\rangle\in\mathcal{U}$.}
    \label{fig:EofF}
    \vspace{-3mm}
\end{figure}
%\end{wrapfigure}
}

% \begin{wrapfigure}{R}{0.45\textwidth}
% %\vspace{-4mm}
%     \begin{minipage}{0.45\textwidth}
       \begin{algorithm}[]
        \caption{{\bf ExpF}}
        \begin{algorithmic}[1]
        \label{alg:EofF}
        \small
        \REQUIRE $\langle X^{(i)},S^{(i)}\rangle,\mathcal{L},C_{t-1}$
        \STATE $sum=0$
        \FOR{$k=0$ to $K-1$}
            \STATE train $C^{i,k}_t$ using $\mathcal{L}\cup\{\langle X^{(i)},S^{(i)},y_k\rangle\}$
            \STATE compute $\mathcal{F}(C^{i,k}_t)$, using $I\mathcal{U}$
            \STATE $sum = sum + \mathcal{F}(C^{i,k}_t) \mathbb{P}(y=k|X^{(i)})$
        \ENDFOR
        \STATE {\bf return} $sum$
        \end{algorithmic}
      \end{algorithm}
%     \end{minipage}
%   \end{wrapfigure}

Since the points to be labeled are selective samples from $\mathcal{U}$, and moved from $\mathcal{U}$ to $\mathcal{L}$, after the process starts, neither $\mathcal{U}$ nor $\mathcal{L}$ can be seen as i.i.d. samples of the actual underlying distribution, and therefore, cannot be used to estimate the fairness.
However, $\mathcal{U}$ initially follows the underlying distribution.
Therefore, to create a dataset for evaluating the fairness of the model for sample selection, we utilize the initial unlabeled pool $\mathcal{U}$ {(referred as $I\mathcal{U}$)} and use it in different FAL iterations.
Following the standard AL, at every iteration, for every possible label for a point $\langle X^{(i)},S^{(i)}\rangle$, SSU uses the current model $C_{t-1}$ for calculating $\mathbb{P}(y=k|X^{(i)})$.
\eat{Having the fairness measures and the probabilities for each possible outcome, the expected unfairness is computed by aggregating $\mathcal{F}(C^{i,k}_t) \mathbb{P}(y=k|X^{(i)})$ for different values of $k$ (Equation~\ref{eq:E[f]}).
After computing the expected unfairness for each data point in $\mathcal{U}$, SSU identifies the one that optimizes Equation~\ref{eq:opt2} and passes it to the labeling oracle.
}
\new{
Algorithm~\ref{alg:EofF} shows the pseudo-code of 
%FAL where Algorithm~\ref{alg:EofF} is invoked 
for computing the expected unfairness.
In order to compute the expected unfairness, Algorithm~\ref{alg:EofF} requires to train $K$ models, each for a possible label for the point $\langle X^{(i)},S^{(i)}\rangle$, which makes it inefficient.
In \S~\ref{sec:fbc}, we propose a {\em constant-time} approximation alternative that enables the \emph{same} asymptotic time complexity as traditional active learning.
We shall then provide the details of extending the fairness block for other measures of fairness beyond DP in \S~\ref{sec:disc}.
}

\new{
Having discussed the FAL framework and the unfairness estimation block, we will next describe the accuracy-fairness optimization block used by SSU for selecting the next sample to be labeled.
}

%% file: tech/opt.tex
\section{Accuracy-Fairness Optimization}\label{sec:alg}
\subsection{FAL $\alpha$-aggregate}\label{sec:opt:alpha}

%We balance the fairness and accuracy trade-off
%by treating fairness as an intrinsic part of the AL sampling strategy.
%The goal is to minimize the misclassification error as well as unfairness (c.f. \S~\ref{sec:back-FM}).
%\techrep{ in order to develop fair classifiers for applications with limited labeled data, similar to Example~\ref{ex1} and Example~\ref{ex2}.}
Similar to AL, FAL is also an iterative process that selects a sample from the unlabeled pool $\mathcal{U}$ to be labeled and added to the labeled pool $\mathcal{L}$.
However, FAL $\alpha$-aggregate considers a combination of unfairness and misclassification error reduction as the optimization objective for the sampling step.
% That is, to choose the next sample to be labeled, FAL $\alpha$-aggregate selects the one that contributes the most to the reduction of the misclassification error as well as model unfairness.
Specifically, for a sample point $\langle X^{(i)},S^{(i)} \rangle\in\mathcal{U}$, we consider the Shannon entropy measure $\mathcal{H}_{t-1}(y^{(i)})$ for misclassification error, while considering demographic disparity $\mathcal{F}(C_{t}^i)$ for unfairness --- $C_{t}^i$ is the classifier trained on $\mathcal{L}$ at iteration $t$, after labeling the point $\langle X^{(i)},S^{(i)}\rangle$
and $\mathcal{H}_{t-1}(y^{(i)})$ is the entropy of the $y^{(i)}$ based on the current model $C_{t-1}$.
\eat{
One way of formulating the optimization problem for sampling is as following:
\vspace{-2mm}\begin{align}
\nonumber
\argmax_{\langle X^{(i)},S^{(i)}\rangle \in\mathcal{U}} ~~&~~ \mathcal{H}_{t-1}(y^{(i)})\\
\nonumber s.t.~~ & ~~\mathcal{F}(C_{t}^i) \leq \epsilon
\end{align}
where $\epsilon$ denotes the restriction imposed by some regulations for fairness consideration.
Similarly, one could consider misclassification error as a hard constraint while optimizing for fairness.
Both of these models can be reformulated as unconstrained optimizations
using \hadis{lagrangian has to be removed or clarified} Lagrange multipliers \cite{rockafellar1993lagrange}.
}
The formulation can be viewed as a multi-objective optimization for fairness and misclassification error.
Another perspective is to view the fairness as a regularization term to the optimization.
Equation~\ref{eq:opt} is consistent with both of these views and is therefore considered in our framework.
% (as in Equation~\ref{eq:opt}) using Lagrange multipliers \cite{rockafellar1993lagrange}.
\vspace{0mm}
\begin{align}
\label{eq:opt}
\argmax_{\langle X^{(i)},S^{(i)}\rangle \in\mathcal{U}}  \alpha\, \mathcal{H}_{t-1}(y^{(i)}) + (1-\alpha)\big(\mathcal{F}(C_{t-1})-\mathcal{F}(C_{t}^i)\big)
\end{align}
$(\mathcal{F}(C_{t-1})-\mathcal{F}(C_{t}^i))$ is the unfairness reduction (fairness improvement) term and the coefficient $\alpha\in[0,1]$ is the user-provided parameter that determines the trade-off between the model fairness and model performance. Values closer to $1$ put greater emphasize on model performance, while smaller values of $\alpha$ put greater importance on fairness.
%Our experimental results verify fairness improves substantially through FAL optimization while maintaining the accuracy level.
As we shall elaborate in \S~\ref{sec:exp}, entropy and fairness values are standardized to the same scale before being combined in Equation~\ref{eq:opt}.

% \subsection{Adaptive $\alpha$ parameter}\label{sec:adaptivealpha}

%S: Slightly changed the writing here as it was duplicating the description of alpha which was introduced in previous para
While $\alpha$ controls the trade-off between the accuracy (entropy) and the fairness terms,
selecting an appropriate $\alpha$ value might not be clear for the user. More importantly, FAL might find it challenging to use a fixed learning strategy based on $\alpha$ in different iterations.
\new{
To avoid parameter tuning in active learning
% In initial iterations, the model is not accurate because it was trained only on a few labeled instances, resulting in possibly inaccurate estimates of the label probabilities for a given unlabeled instance from $\mathcal{U}$.
% Computing the expected fairness values relies heavily on the probabilities of the label. The miscalculation of these probabilities leads to an inaccurate estimation of fairness; such erroneous values contribute to the selection of points that do not support (and may even deteriorate) the fairness of the model and may not be good for model accuracy.
% %Wrong samples selected due to the wrong estimation of unfairness may not be a good choice, either from accuracy nor fairness perspective.
% In the later iterations, the model may already be stable and accurate, and new labeled points may not significantly impact its accuracy.
% However, the model can provide better estimations of the label probabilities which results in more robust estimations of the expected fairness.
% %The points that are introduced in the later iterations may not significantly impact the model accuracy, and yet can have a better impact on the model fairness improvement.
instead of using a fixed $\alpha$ for all iterations, one can use a {\em decay function} that begins with a large value of $\alpha$, which improves the accuracy of the model. As the model becomes more stable, the value of $\alpha$ gets dropped, putting more weight on improving fairness.
This concept is applied in different context, such as assigning learning rate~\cite{schaul2013no}, where a larger value is used initially that gradually decreases over time.
Our approach is agnostic to the decay function used.
In the experiments, we use a function that linearly interpolates between the range [0,1].
}
\new{
The pseudo-code of FAL with $\alpha$-aggregate is provided in Algorithm~\ref{alg:fal}.
}

\begin{algorithm}[!t]
    \caption{FAL $\alpha$-aggregate}
    \begin{algorithmic}[1]
    \label{alg:fal}
    \small
    \FOR{$t=1$ to $B$}
        \STATE $max=0$
        \FOR{$i=1$ to $|\mathcal{U}|$}
            \STATE {\it H} =$-\sum_{k=0}^{K-1} \mathbb{P}(y=k|X)\log \mathbb{P}(y=k|X)$
            \STATE $F =$ {\bf ExpF}$\big(\langle X^{(i)},S^{(i)}\rangle,\mathcal{L},C_{t-1}\big)$
            \STATE $obj = \alpha(i) H + (1-\alpha(i))(\mathcal{F}(C_{t-1})-F)$
            \IF{$obj>max$}
                \STATE $max = obj$;
                $\langle X^*,S^*\rangle = \langle X^{(i)},S^{(i)}\rangle$
            \ENDIF
        \ENDFOR
        \STATE $y =$ label $X^*$ using the labeling oracle
        \STATE move $\langle X^*,S^*,y\rangle$ to $\mathcal{L}$
        \STATE train the classifier $C_t$ using $\mathcal{L}$
    \ENDFOR
    \STATE {\bf return} $C_t$
    \end{algorithmic}
\end{algorithm}

\subsection{FAL Nested}\label{sec:opt:nested}
Accurately estimating the expected unfairness reduction is critical for the performance of FAL.
Looking at Equation~\ref{eq:E[f]}, computing the expected fairness directly depends on the probability estimations of the current model $C_{t-1}$ for different class labels $k$. The miscalculation of these probabilities leads to an inaccurate estimation of fairness.
Sacrificing accuracy for fairness in previous steps will affect the estimation of expected fairness in subsequent steps. Such erroneous estimation of the expected unfairness improvement as a result, contribute to the selection of points that do not support (and may even deteriorate) the fairness of the model and may not be good for model accuracy.
%may mislead the sample selection to choose points that are not only not good for accuracy, but also not good for improving fairness.
In order to prevent this phenomena, it is necessary to always maintain an accurate intermediate model -- i.e., to ensure that the selected points to be labeled will not negatively impact the accuracy of the model.

Fortunately, we make an observation in practice that helps us keeping the intermediate models accurate while achieving fairness.
As we shall further explain in the following, our observation, also helps us to even reduce the computation cost of the model by only focusing on a subset of $\mathcal{U}$, instead of the full set.

It turns out in practice the distribution of the entropy of the data points in $\mathcal{U}$ is right-skewed, i.e. having a large number of points with entropy close to 1, i.e., all these points are almost equally good candidates from the accuracy perspective. We observed this in our real experiments, including the one on COMPAS dataset shown in Figure~\ref{fig:edist}.

Consider the set of points on the right-most bucket with entropy close to one.
Any of the points in this set are good candidates from lens of active learning (Equation~\ref{eq:entropy}) to ensure the high quality of the intermediate models.
On the other hand, the relatively large size of the set (as observed in Figure~\ref{fig:edist}) increases the chance that it contains a point with a high potential for reducing unfairness.

\eat{
Note that the points with largest entropy may not be the same points with largest unfairness reduction. Hence, when we combine the entropy and fairness metric with the control parameter $\alpha$, the distribution of the score based on Equation~\ref{eq:opt} will be less skewed as we decrease $\alpha$. This allows SSU selects the points that are not improving the accuracy necessarily (the points that does not belong the largest bucket on $\mathcal{H}$ distribution). We would like to underscore that the high accuracy of the model after adding the selected candidate point is critical to accurately estimating the expected (un)fairness reduction values in the subsequent iterations.
Hence, sample selection based on the combined score, Equation~\ref{eq:opt}, in such situation may result in a less accurate model, and misleading the SSU in choosing the points that will improve the (un)fairness reduction.
In this section, we propose a hierarchical sampling for accuracy-fairness trade-off to address this issue and prevent the random selection.
}

We use the above observation to design a {\em nested optimization} for accuracy and fairness.
In particular, instead of computing a score by linearly combining the two terms (Equation~\ref{eq:opt}), we apply a nested optimization where in the first level, we select a subset of $\mathcal{U}_A \subset \mathcal{U}$ of the top-$\ell$ points\footnote{Instead of a fixed set cardinality, one could consider selecting the points that their values are close to the maximum (e.g. have distance less than 0.001).} that maximize the entropy $\mathcal{H}(y^{(i)})$ using the current classifier $C_{t-1}$:

\begin{align}\label{eq:top-ell}
    \mathcal{U}_A = \underset{\langle X^{(i)},S^{(i)}\rangle \in\mathcal{U}}{\ell\mbox{-argmax}} \mathcal{H}(y|X,\mathcal{L})
\end{align}

where the function $\ell$-argmax returns $\ell$ samples with maximum values.
The first optimization level only optimizes for accuracy, to ensure maintaining a high accuracy for the intermediate model.
%The next level optimization
Next, changing the optimization criteria to fairness in the second level, SSU selects a point from $\mathcal{U}_A$ that maximizes the unfairness improvement (Equation~\ref{eq:top-kp2}) and pass it to the labeling oracle.
\begin{align}
    \nonumber&\argmax_{\langle X^{(i)},S^{(i)}\rangle \in\mathcal{U}_A} \big(\mathcal{F}(C_{t-1})-\mathcal{F}(C_{t}^i)\big)\\
    =& \argmin_{\langle X^{(i)},S^{(i)}\rangle \in\mathcal{U}_A} \mathcal{F}(C_{t}^i)
    \label{eq:top-kp2}
\end{align}

\eat{
Consider $\mathcal{U}_A\subset \mathcal{U}$ a set of cardinality $\ell$ maximizing entropy $\mathcal{H}(y^{(i)})$ using the current classifier $C_{t-1}$. $\forall P_i \in \mathcal{U}_A$ we evaluate the expected (un)fairness reduction $E\big[\mathcal{F}^{i}_{t}\big]$. The points with the maximum $E\big[\mathcal{F}^{i}_{t}\big]$ value will be selected to be labeled next. When $\ell=1$ FAL top-$\ell$ performs similar to active learning itself. The smaller the $\ell$ is the more weight the SSU puts on accuracy.
As $\ell$ increases SSU will have access to more points that improves the expected unfairness reduction.
}

Besides assuring the accurate estimations of expected unfairness reduction, the nested optimization helps to reduce the time-complexity of FAL.
This is especially important as it reduces the number of points to be evaluated for fairness from $|\mathcal{U}|$ to $\ell$.
Note that, in order to estimate the unfairness reduction for every sample (Figure~\ref{fig:EofF}), Algorithm~\ref{alg:EofF} requires retraining of $k$ models. As a result, the computation time to run the framework is dominated by the total time to compute fairness values for the points in the unlabeled pool, before selecting one to be labeled.
Hence, reducing the number of points to be evaluated for fairness significantly reduces the computation cost by a factor of $\frac{\ell}{|\mathcal{U}|}$.

\eat{%condense
\begin{figure}[!t]
\centering
    %\vspace{-3mm}
    \includegraphics[width=\linewidth]{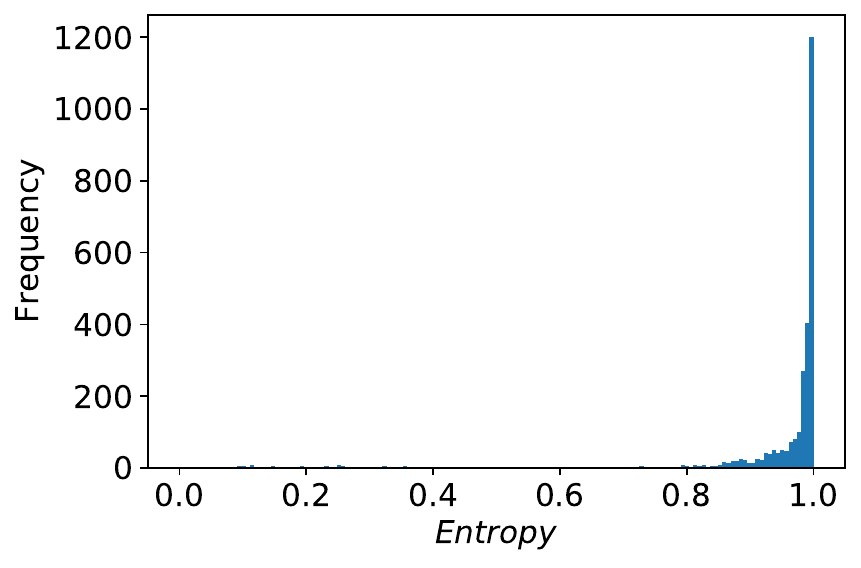}
    \vspace{-7mm}
    \caption{\small Distribution of Entropy for COMPAS dataset.}
    \label{fig:edist}
    \vspace{-5mm}
%\end{wrapfigure}
\end{figure}
}

% It is worth noting that the computation time of FAL top-$\ell$ approach significantly improves compare to FAL since we the expected (un)fairness reduction is only evaluated for the restricted dataset $\mathcal{U}_{|A|}$ with size $\ell$.

\subsection{FAL Nested-Append}\label{sec:opt:nAppend}
Since the sample points in $\mathcal{U}$ are unlabeled, FAL has no choice but to estimate the expected unfairness reduction after labeling a point, according to how likely it will take each possible label.
But whether unfairness reduces depends on the actual label after adding the point to $\mathcal{L}$.
To further clarify this, let us consider a toy example highlighted in Figure~\ref{fig:append}.
To simplify the explanation, we use 6 triangle and 6 circle points, each representing a demographic group, to evaluate the fairness of a linear binary classifier.
Suppose the decision boundary of current classifier ($C_{t-1}$) is the one shown in solid line in the figure, and the points in the top-right of the line are classified as +1.
$C_{t-1}$ classifies two third (4 out of 6) of circles, but only one third (2 out of 6) of triangles as +1, hence is not fair according to DP.

\eat{%condense
\begin{figure}[!ht]
\centering
    %\vspace{-3mm}
    \includegraphics[width=0.7\linewidth]{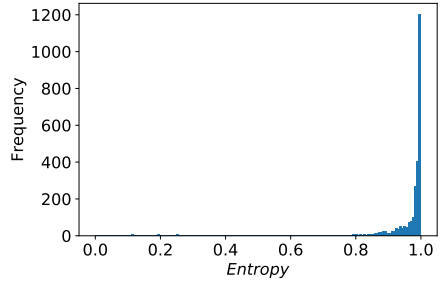}
    \vspace{-5mm}
    \caption{\small Distribution of Entropy for COMPAS dataset.}
    \label{fig:append}
    \vspace{-4mm}
%\end{wrapfigure}
\end{figure}
}

Looking at the figure, to make the classifier fair, FAL needs to rotate the border towards the dashed line.
Consider the two angles highlighted in the figure in the intersection of the two lines.
To make the rotation, FAL needs to find points in $\mathcal{U}$ with the true +1 label
that belong to the top-left angle, or the ones with the true -1 label in the bottom-right angle.
Note that such points are misclassified by the current classifier. As a result, it is less likely to find the points with proper labels needed for reducing unfairness.
On the other hand, if the label is not as hoped, adding the new point will not help to reduce unfairness.
Therefore, to boost FAL for improving fairness,
our next optimization, FAL Nested-Append, replicates the points that get labeled in a way that unfairness gets reduced.
Let $\Tilde{P}_i=\langle \Tilde{X}^{(i)},\Tilde{S}^{(i)}\rangle \in \mathcal{U}_A$ be the point selected using FAL Nested.
Let $\mathcal{F}(C_t)$ be the unfairness of the model after collecting the true label of $\Tilde{P}_i$ and adding it to $\mathcal{L}$.
If $\mathcal{F}(C_t)-\mathcal{F}(C_{t-1})>0$, the algorithm replicates $\Tilde{P}_i$ in $\mathcal{L}$, further boosting its impact for unfairness reduction.
In particular, since {FAL Nested} puts more emphasize on accuracy, {FAL Nested-Append} helps to account more for fairness.
% In this way, the algorithm assigns a weight to every point according to its contribution to the unfairness reduction.
% Of course, one could have used other weighting strategies, and our algorithm is agnostic to such choices.
As we shall show in \S~\ref{sec:exp}, boosting the performance of {FAL Nested} for fairness, {FAL Nested-Append} had the {\em best performance} across different experiments.

\eat{
Although in FAL Top-$\ell$ approach SSU selects points with maximum expected (un)fairness reduction from the top-$\ell$ maximum entropy points, the selected candidate might not necessarily improve the true (un)fairness reduction due to the low overall accuracy of the current classifier (especially in the initial iterations).
To further improve the true (un)fairness level of the classifier, we propose to replicate the points that actually improve the true (un)fairness reduction after it is labeled. In this way we assign a weight to every point according to its contribution to the true (un)fairness reduction.

Let $\Tilde{P}_i=\langle \Tilde{X}^{(i)},\Tilde{S}^{(i)}\rangle \in \mathcal{U}_A$ be the selected point using FAL top-$\ell$ approach. After collecting the true label for $\Tilde{P}_i$, let $\Tilde{C}^{i,k}_t$ be the model and $\Tilde{\mathcal{F}}^{i}_{t}$ be the true (un)fairness of the model after adding $\Tilde{P}_i$ to $\mathcal{L}$ if its true label is $\Tilde{y}^{(i)}=k$. $\langle \Tilde{X}^{(i)},\Tilde{S}^{(i)}\rangle,k$ will be replicated once if $\mathcal{F}^{i}_{t-1}-\Tilde{\mathcal{F}}^{i}_{t} > 0$.
}

%% file: tech/fbc.tex
% \vspace{-5mm}
\section{Efficient FAL by Covariance}\label{sec:fbc}
Our proposed FAL framework is agnostic to the choice of classifier $C$.
In this section, we show that it is possible to design efficient
algorithm for the special case of generalized linear models.
An appealing property of our algorithm is that, \new{using the efficient computation provided in Appendix~\ref{app:FBC-m},} it has
\emph{same} asymptotic time complexity as traditional active learning.
%As an efficient alternative for fair active learning, in this section we propose an alternative method for achieving fairness for generalized linear models
We achieve this by avoiding the model retraining step for calculating the expected fairness of unlabeled samples, in Algorithm~\ref{alg:EofF}.

Consider a generalized linear model in form of $\hat{y}=\theta^\top X$.\footnote{The decision boundary of the classifier is viewed as a threshold value on $\hat{y}$ that separate different classes, using, for example, a sign function.} %\hadis{if y hat is a classifier it needs a sign or sigmoid function.}
The covariance between the model and a sensitive attribute $S$ should be zero under model independence (demographic parity).
We make a key observation in Lemma~\ref{th:cov} that shows this covariance, $cov(S,\hat{y})$, {\em only depends on $cov(S,X)$ and $\theta$}.
% In other words, the underlying covariance $cov(S,X)$ can cause model unfairness.

\begin{lemma}\label{th:cov}
For a generalized linear model in form of $\hat{y}=\theta^\top X$,
% Under the Assumption 1.
$cov(S,\hat{y})=\theta^\top cov(S,X)$.\footnote{Please refer to Appendix~\ref{app:proof1} for the proof of Lemma~\ref{th:cov}.}
% \footnote{The proof of Lemma~\ref{th:cov} can be found in the \techrep{Appendix}\submit{supplementary material}.}
\end{lemma}

According to Lemma~\ref{th:cov}, the covariance of the model with the sensitive attribute (that results in unfairness) depends only on the weight vector $\theta$ and the underlying covariance of features $X$ with $S$.
We can reduce the model unfairness by ensuring that the model does not assign high weights to the problematic features (the features with high covariance with $S$).
This observation allows us to indirectly optimize for fairness through covariance instead of computing expected unfairness reduction.

% for a specific tuple and a feature
Consider a feature $x_i$ that is highly correlated with the sensitive attribute (i.e., $cov(x_i,S)$ is high) and also has a high weight $\theta_i$ in the current model. Our objective is to reduce the weight assigned to such features.
The reason the model has assigned a large weight to $x_i$ is that {\em $x_i$ is highly predictive of $y$ in $\mathcal{L}$}. %In other words, $x_i$ is a good signal for predicting $y$ according to $\mathcal{L}$.
% The high covariance between $x_i$ and $y$ in $\mathcal{L}$ has made it a valuable signal for the model to predicting $y$.
Therefore, in order to reduce the weight $\theta_i$, we need to reduce $cov_{\mathcal{L}}(x_i,y)$ in the labeled pool $\mathcal{L}$ to make it less predictive of $y$ in $\mathcal{L}$.
Now, consider a point $P_j = \langle X^{(j)}, S^{(j)} \rangle\in \mathcal{U}$ and its value $X^{(j)}_i$ on feature $x_i$.
Depending on $X^{(j)}_i$ and its label $y^{(j)}$ (after labeling), the point $P_j$ can impact $cov(x_i,y)$ in $\mathcal{L}$.
Indeed, we don't know $y^{(j)}$ during the sample selection step.
Still, similar to \S~\ref{sec:alg}, we can consider the probability distribution over $y$ and calculate the expected improvement in covariance.
%treat the covariance improvement as expected value.
Let $cov_i = cov_{\mathcal{L}}(x_i,y)$ be the covariance of $x_i$ and $y$ in $\mathcal{L}$
and $cov_{j,i,k} = cov_{\mathcal{L}\cup\{\langle X^{(j)}, S^{(j)},k \rangle\}}(x_i,y)$
the covariance of $x_i$ and $y$ after adding $\langle X^{(j)}, S^{(j)},k \rangle$ to $\mathcal{L}$.
The expected covariance improvement for $x_i$ after adding $P_j$ to $\mathcal{L}$ is
%\vspace{-2mm}
\begin{align} \label{eq:efibc}
E\big[cov^{\downarrow}_{j,i} \big]  = \sum_{k=0}^{K-1} (|cov_i| - |cov_{j,i,k}|) \mathbb{P}(y=k|X^{(j)})
\end{align}
Following Lemma~\ref{th:cov} the contribution of the covariance reduction for a feature $x_i$ to fairness is proportional to $|\theta_i cov(x_i,S)|$.
Subsequently, it is important to reduce $cov_{\mathcal{L}}(x_i,y)$ for the features that are highly correlated with the sensitive attribute and have a high weight $\theta_i$ in the model.
Therefore, the (indirect) fairness improvement by covariance for a point $P_j\in\mathcal{U}$ can be computed as following:
%\vspace{-2mm}
\begin{align}\label{eq:efbc}
    E\big[FbC^{\downarrow}_{j} \big] =
    \sum\limits_{i=1}^d |\theta_i cov(S,x_i)| E\big[cov^{\downarrow}_{j,i} \big]
\end{align}
Now, it is enough to replace the term for expected unfairness reduction $\big(\mathcal{F}(C_{t-1})-\mathcal{F}(C_{t}^i)\big)$, in Equations~\ref{eq:opt} and~\ref{eq:top-kp2}, with $E\big[FbC^{\downarrow}_{i} \big]$.
%In \submit{the supplementary material}\techrep{the Appendix},

%% file: tech/otherfairness.tex
\section{Extension to Other Fairness Models}\label{sec:disc}
So far in this paper,
we considered  independence ($\hat{y}\bot S$) for fairness.
Next, we discuss how to extended our findings to other measures based on separation and sufficiency \cite{fairmlbook}, such as equalized odds \cite{hardt2016equality}, where the prediction outcome $\hat y$ is independent of the sensitive feature $S$ given the true label $y$, i.e. $P(\hat{y}=1|y = 1, S = 0)
= P(\hat{y}=1|y = 1, S = 1), y \in {0,1}$.

% predictive parity, error rate balance, and accuracy equity~\cite{narayanan2018translation}.
\new{
The fairness-accuracy optimizer of FAL is not limited to a specific notion of fairness for balancing accuracy and fairness. Similarly, the notion of expected unfairness reduction does not rely on a specific notion of fairness as $\mathcal{F}(.)$ in Equation~\ref{eq:E[f]} can be computed using any fairness measure, besides demographic parity.
As a result, at a high-level, the FAL framework should work as-is for other notions of fairness as well.
% using Equation~\ref{eq:opt}. Certainly, the entropy term $\mathcal{H}_{t-1}(y^{(i)})$ does not depend on the choice of fairness measure.
% Also, the abstract fairness term $E\big[\mathcal{F}(C_{t}^i(.))\big]$ is not limited to a specific definition.
However, as we shall explain in the following, computing $\mathcal{F}(.)$ would require additional information that comes at a cost of randomly labeling a subset of data.
}

% despite being abstract, computing the expected unfairness (based on separation or sufficiency) is challenging.

Looking at Figure~\ref{fig:EofF}, recall that we use the initial unlabeled set $I\mathcal{U}$ for estimating the fairness of a model.
$I\mathcal{U}$ follows the underlying data distribution and, hence, can be used for evaluating the demographic disparity.
However, this set cannot be used for estimating fairness according to separation or sufficiency since its instances are not labeled.
On the other hand, the pool of labeled data are not representative of the underlying data distribution.

\new{
Our resolution is to use a small subset of $\mathcal{C}_\mathcal{U}\subset I\mathcal{U}$ for fairness computation, accepting the potential error in estimations relative to the size of the set.
% In order to extend our results for other fairness measures, it is enough to label a random subset $\mathcal{C}_\mathcal{V}\subset\mathcal{V}$.
Before starting the FAL process, we need to label $\mathcal{C}_\mathcal{U}$.
Once labeled,
$\mathcal{C}_\mathcal{U}$ will be used for calculating $\mathcal{F}(.)$ based on other notions of fairness, and
FAL can be executed as-is.
In \S~\ref{sec:exp}, we run experiments to show the extension of FAL for equalized odds.
}
% How small the subset can be to still provide accurate-enough estimations, besides other resolutions, are interesting questions that we will consider for future work.

%% file: results.tex
\vspace{-3mm}
\section{Experiments}\label{sec:exp}
% \subsection{Experiments Setup}\label{sec:setup}
The experiments were performed on a Linux machine with a Core I9 CPU and 128GB memory. The algorithms were implemented using Python 3.7
\footnote{Our codes are publicly available: \url{https://github.com/anahideh/FAL--Fair-Active-Learning}}.

\vspace{-3mm}
\subsection{Datasets}\label{sec:exp:datasets}
% We apply the proposed Fair Active Learning approach to three different datasets.

{\it COMPAS}\footnote{ProPublica, \url{https://bit.ly/35pzGFj}}: published by ProPublica \cite{propublica}, this dataset contains information of juvenile felonies
such as \attrib{marriage status, race, age, prior convictions}, and the \attrib{charge degree} of the current arrest.
We normalized data so that it has zero mean and unit variance.
We consider \attrib{race} as sensitive attribute and filtered dataset to black and white defendants. The dataset contains 5,875 records, after filtering.
Following the standard practice~\cite{corbett2017algorithmic,mehrabi2019survey}, we use two-year violent recidivism record as the true label of recidivism: $y^{(i)}=1$ if the recidivism is greater than zero and $y^{(i)}=0$
otherwise.

\noindent{\it Adult dataset}\footnote{UCI repository, \url{https://bit.ly/2GTWz9Z}}: contains 45,222 individuals income extracted from the 1994 census data with attributes such as \attrib{age, occupation, education, race, sex, marital-status, hours-per-week,} \attrib{native country}, etc. We use \attrib{income} (a binary attribute with values $\geq 50k$\$ and $\leq 50k$\$) as the true label. We consider \attrib{sex} as the sensitive attribute. We normalized data so that it has zero mean and unit variance.

%Both of these datasets are widely used for evaluating the design of fair classifiers.
% In the experiments, we do not recognize the sensitive attributes as model features, so as not to practice {\em disparate treatment}.

\eat{
{\it German Credit Score}\footnote{UCI repository, \url{https://bit.ly/36x9t8o}}: includes 1000 individuals credit records containing attributes such as marriage status, sex, credit history, employment, and housing status. We normalized data for the inputs and outputs to be mean zero and unit variance. We consider \attrib{sex} as the sensitive attribute. We use credit rating (0 for bad customers and 1 for good customers) as the true label, $y^{(i)}$, for each individual.
 }

% {\it Adult}:
% \input{plots-new.tex}
\input{plots-new}

\subsection{Algorithms Evaluated}
%We evaluate the performance of the following approaches on the benchmark datasets in \S\ref{sec:exp:datasets}.
All our proposed approaches are evaluated using a
regularized $\ell_2$ norm logistic regression classifier with a regularization strength of one.
%as the classifier in all of the cases.
We trained the logistic regression with \emph{liblinear} optimizer and with maximum iteration of 100.
Our findings are transferable to other classifiers.
We begin by comparing our proposed approach against a wide variety of representative baselines.
Then, we focus on understanding the effectiveness and performances
of our proposed approaches under different settings.
%\subsection{Adaptive $\alpha$}

\stitle{Baselines:}
We consider four baselines in order to build a fair classifier in a limited data environment. We first start to evaluate passive methods, {\bf RandL} and {\bf R-FLR}, that select all the samples randomly at one shot to form a training set and fit a regular and fair logistic regression (proposed by ~\cite{zafar2017fairness}), respectively. We then evaluate active methods, {\bf AL} and {\bf AL-FLR}, which iteratively select a sample point based on its informativeness (Equation~\ref{eq:entropy}) and fits a regular and fair logistic regression ~\cite{zafar2017fairness} in each iteration, respectively.

\stitle{Our Algorithms:}
We evaluate our fairness-accuracy optimization algorithms proposed in \S~\ref{sec:alg}, namely FAL $\alpha$-aggregate ({\bf FAL-$\alpha$}), FAL Nested ({\bf Nested}), and FAL Nested-Append ({\bf N-App}).
For {\bf FAL-$\alpha$}, we normalize the accuracy and fairness improvement values as ($v$-min)/(max-min) before combining them in Equation~\ref{eq:opt}.
Besides fixed values of $\alpha$, we also consider an adaptive $\alpha$ parameter, using a decay function that, starting from $\alpha=1$ to $\alpha=0$, the value $\alpha$ drops by 0.1 every $\lfloor B/11 \rfloor$ iterations. For {\bf Nested}, and {\bf N-App}, we consider different exponents of 2 as the value of $\ell$ (in Equation~\ref{eq:top-ell}) from $2$ to $512$. 
\eat{
We use the comparison between {\bf FAL-$\alpha$}
We also evaluate the performance of FAL Nested ({\bf Nested}) approach and FAL Nested-Append ({\bf N-App}) to show the significant improvement on the efficiency and effectiveness of our proposed {\bf FAL-$\alpha$} approach.
}

Our default choice for computing unfairness reduction is Algorithm~\ref{alg:EofF}.
The efficient FAL by covariance ({\bf FBC}), proposed in \S\ref{sec:fbc}, is also evaluated to show the computation time improvement. We evaluate {\bf FBC} with the three optimization approaches {\bf FAL-$\alpha$}, {\bf Nested}, and {\bf N-App}.
Finally, in order to show the extension of our proposal for other fairness models, we run FAL using Equalized Odds as the fairness measure.

\vspace{-4mm}
\subsection{Performance Evaluation}\label{sec:performance}
We perform the experiments using 30 random splits of the datasets into training $\mathcal{U}$ ($60\%$ of the examples) and testing ($40\%$ of the examples).
%We study the values of $\{0.6,0.8\}$ for $\alpha$, plus the adaptive $\alpha$ approach.
We consider the mean and variance over the 30 random splits.
We specify the maximum labeling budget to 200 after which performance leveled off.
%, where the performance leveled off in our preliminary results.
In each FAL and AL scenarios, we start with six labeled points and sequentially select points to label, until the budget is exhausted.
Mutual information is our default measure of demographic parity.
%We also use the Absolute Difference $|\mathbb{P}(\hat{y}=1|S=0) - \mathbb{P}(\hat{y}=1|S=1)|$ as another metric for measuring DP.

We first evaluate the performance of  \textbf{FAL-$\alpha$} versus the passive and active baselines \textbf{RandL}, \textbf{R-FLR}, \textbf{AL}, and \textbf{AL-FLR}, using accuracy and fairness measures to show the deficiency of these approaches in construction of a final fair classifier.
Figure \ref{exp:base} illustrates the performance of baselines and \textbf{FAL-$\alpha$} where $\alpha=0.6$ for COMPAS and Adult datasets.
%using $\{0.6\}$ for $\alpha$,
The bars indicate the standard deviation on 30 random split of data. We observe that the baselines had similar performances on fairness.
Even applying a fair classifier ({\bf FLR}) fails to improve the fairness.
{\bf FAL-$\alpha$}, on the other hand, significantly reduces unfairness while sustaining a comparable accuracy.

% dramatically mitigates the unfairness measure in both cases compared to baselines while sustaining a comparable accuracy.

In our next experiment illustrated in Figure~\ref{exp:fal}, we evaluate the average performance of three different optimizers {\bf FAL-$\alpha$, Nested}, and {\bf N-App}, the efficient algorithms proposed in \S\ref{sec:alg} and compare it against AL.
Figure \ref{exp:alpha-compas} presents a comprehensive comparison of FAL with different user-defined $\alpha$ and adaptive alpha, versus AL on COMPAS dataset. We can observe that {\bf FAL-$\alpha$} achieved a good level of fairness across different $\alpha$ values. Figure \ref{exp:nested-compas} corresponds to the performance of {\bf Nested}, which focuses on the upper percentile of the entropy distribution, to ensure that the selected points are improving accuracy, and not only the unfairness. As expected, the results indicate that this approach nudges up the accuracy of {\bf FAL-$\alpha$}.
Finally, in Figure~\ref{exp:append-compas}, we evaluate the average performance of {\bf N-App} on COMPAS dataset. Compared to both  {\bf FAL-$\alpha$} and {\bf Nested}, the unfairness level of the model dramatically improved by appending the points two times when they truly improve the unfairness level in each iteration.
Similarly, we replicated the results for Adult dataset as in Figure~\ref{exp:alpha-adult}, \ref{exp:nested-adult}, and \ref{exp:append-adult}. It can be seen that the effective {\bf N-App} approach significantly improves the unfairness level of the model while maintaining its accuracy.

We also evaluate the average performance of {\bf FBC} approach as proposed in \S\ref{sec:fbc}. Figure \ref{exp:fbc} includes the results of {\bf FBC} with the three proposed optimizers on COMPAS dataset (the results for Adult dataset are provided in Appendix~\ref{app:exp}). The results are fairly consistent with the results we observed in Figure \ref{exp:fal} for COMPAS dataset. Note that {\bf FBC} is an approximation of the expected unfairness reduction and is computationally more efficient to be used in FAL algorithm (Figure~\ref{exp:time}).
% \hadis{ I am here}
% We can observe that FAL achieved a good level of fairness by mitigating the bias while collecting the training data points, and applying fair LR on the labeled samples help to further improve fairness.
% However, the impact of fair LR is not pronounced. This is due the fact that the sample generated by FAL is already unbiased thereby achieving a fair classifier even when trained using a traditional classification algorithm.
As we discussed in \S\ref{sec:disc}, our proposed approaches can be extended to use other fairness measures.
Figure \ref{exp:odds-compas} corresponds to the experimental setup where he Equalized Odds notion of fairness as proposed in \cite{hardt2016equality} is used in {\bf FAL-$\alpha$} for measuring fairness. The results indicate the effectiveness of FAL compared to AL using different $\alpha$ values. The results of the same setting for Adult dataset is provided in Figure~\ref{exp:odds-adult} in Appendix.
% Note that the results will significantly be improve using the augmented optimizers {\bf Nested} and {\bf N-App}

% However, FAL with user-defined $\alpha$ slightly outperforms the adaptive idea from fairness perspective.

Figure \ref{exp:budget} corresponds to the average (un)fairness and average accuracy score on 30 random runs. Looking at the figure, we can observe that {\bf Nested-64} enforces the accuracy while considering the fairness in the sample selection. Hence, with a higher accuracy and lower unfairness it outperforms {\bf FAL-$\alpha$-0.6}. Note that the performance of {\bf N-App-64}, outperforms both {\bf FAL-$\alpha$-0.6} and {\bf Nested-64} as expected.

Figure~\ref{exp:time} shows the computation time of each sampling iteration for different accuracy-fairness optimizers compared to the original FAL.
% Figures~\ref{exp:fal-fbc-demo} and \ref{exp:fal-fbc-score}, show the average fairness and accuracy values.
FBC is orders of magnitude faster than FAL as it avoids the need to compute expected fairness. On the other hand, since it indirectly optimizes for fairness, FAL outperforms it on fairness.

% In summary, the results of the experiment support the effectiveness of our proposal as FAL could significantly improve { \it fairness } by reducing the disparities while the model {\it accuracy} (fraction of correct predictions) does not significantly impact the overall accuracy of the model across different $\alpha$ values.

\eat{Figures~\ref{exp:demo11} and \ref{exp:pr11} corresponds to employing Mutual Information, $Fairness_{11}$, in optimization step. We note that adding the fairness measure to the optimization for candidate selection does not significantly impact the overall accuracy of the model across different $\alpha$ values, Figure~\ref{exp:demo11}. Compared to AL also, the accuracy does not drop dramatically. As we observe, the demographic disparity of the model though, dropped by almost $50\%$ on average compared to AL and RandL.} %, especially when the weight of the fairness is higher, $\alpha=0.6$.
\eat{Figure~\ref{exp:pr11} indicates that AL and RandL have fairly lower precision and recall. Precision and recall are better evaluation measure for unbalanced data classification, thereby, FAL outperforms AL and RandL in predicting the right class.}

% Moving to second fairness measure, Figures~\ref{exp:demo12} and \ref{exp:pr12} shows the results using $Fairness_{12}$, the covariance metric, as fairness metric in optimization. We note that the accuracy of the model across different scenarios has a negligible difference. The covariance metric of FAL on average is lower than AL and RandL.

% Figures~\ref{exp:accu21}--\ref{exp:prec21} provide the results where we use demographic error measure $Fairness_{21}$ for the optimization fairness component. Here, there is an insignificant change in the accuracy score as the $\alpha$ value changes. We can observe the same overall increasing pattern for the fairness metric with the best fairness value occurring at $\alpha=0.6$. It is worth mentioning that, the F1-score and precision/recall have their best value compared to other scenarios with different fairness metrics.

\eat{Employing $Fairness_{22}$ in the optimization process leads to the results presented in Figures~\ref{exp:demo22} and \ref{exp:pr22}. A very small change can be seen in the accuracy score through different values of $\alpha$ and the same pattern of fairness is observable.}

\eat{Results provided in Figures~\ref{exp:demo31} and \ref{exp:pr31} correspond to the $Fairness_{31}$ metric applied in optimization process. The results are consistent with our findings in previous scenarios; the fairness measure is improved in FAL across different $\alpha$ values, whereas the accuracy does not change, remarkably.}

% Figures~\ref{exp:accu32}--\ref{exp:prec32} presents the results for the case $Fairness{32}$ is applied as the fairness metric in optimization. Reasonably stable accuracy with an increasing fairness trend can be noticed. There is an F1-score drop at $\alpha=0.7$, which is resulted from the precision drop under that scenario.

% Next, we study the trend of accuracy and fairness as the labeled instances increases under different fairness scenario. We label each FAL scenario as FAL-$Fairness_{ij}$. We provide the results for a single run at $\alpha=0.6$. Each pair in figures~\ref{exp:b:accu11}--\ref{exp:b:demo32} corresponds to the accuracy and fairness measure, respectively.
% Overall, as the number of sample points increase the accuracy of the model increases. Also, the fairness component tries to maintain the model fairness over time which is observed in our results.
% Initially, the models are not reliable as the number of training samples were small. This resulted in an unstable behaviour of the models both for accuracy and fairness in initial steps. The trends become more stable after a few iterations that models become more reliable.

%% file: plots-new.tex
\begin{figure*}[!t]
\centering     %%% not \center
\subfigure[FAL $\alpha$-aggregate]{\label{exp:alpha-compas}\includegraphics[width=0.32\linewidth]{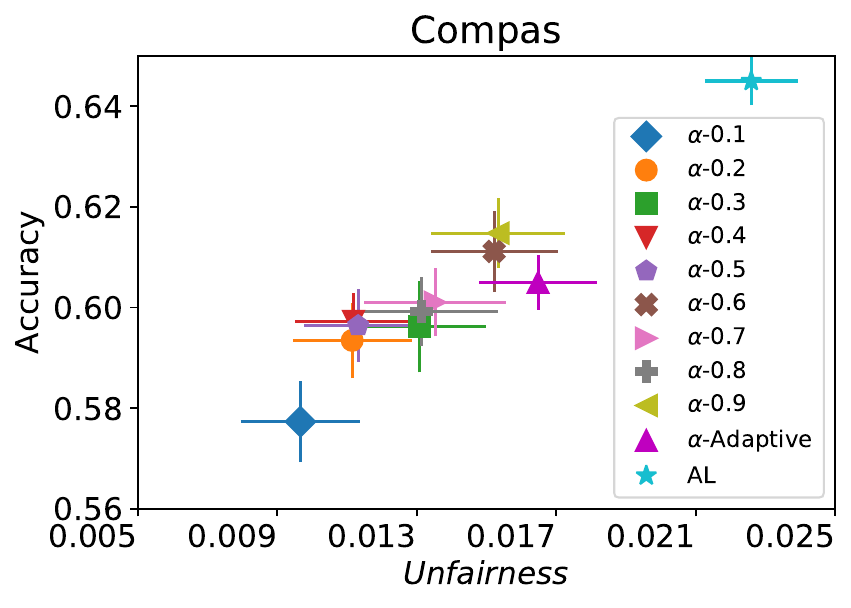}}
\subfigure[FAL Nested]{\label{exp:nested-compas}\includegraphics[width=0.32\linewidth]{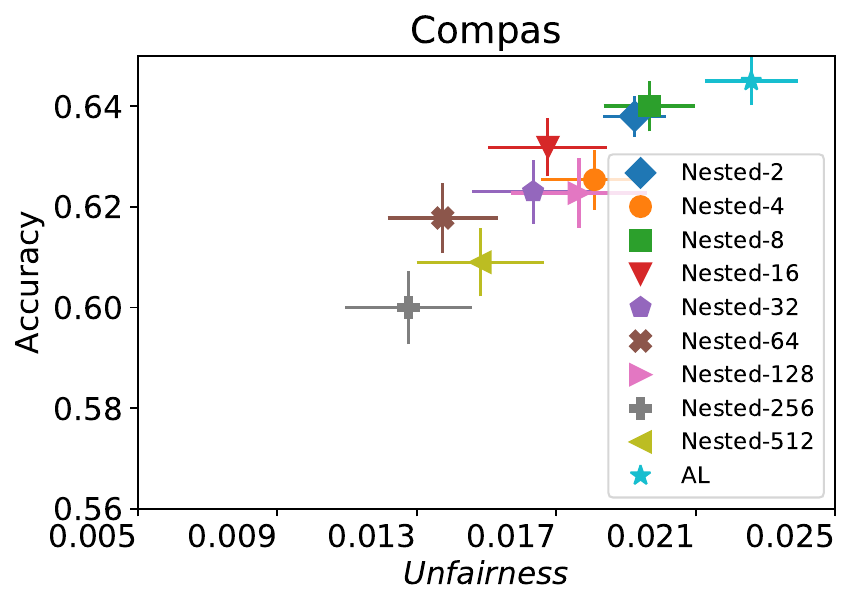}}
\subfigure[FAL Nested-Append]{\label{exp:append-compas}\includegraphics[width=0.32\linewidth]{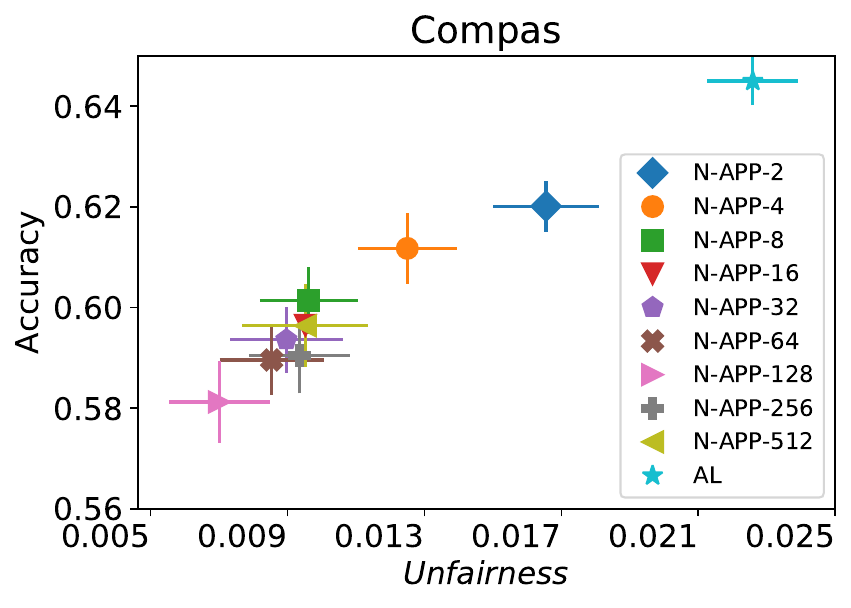}}
\subfigure[FAL $\alpha$-aggregate ]{\label{exp:alpha-adult}\includegraphics[width=0.32\linewidth]{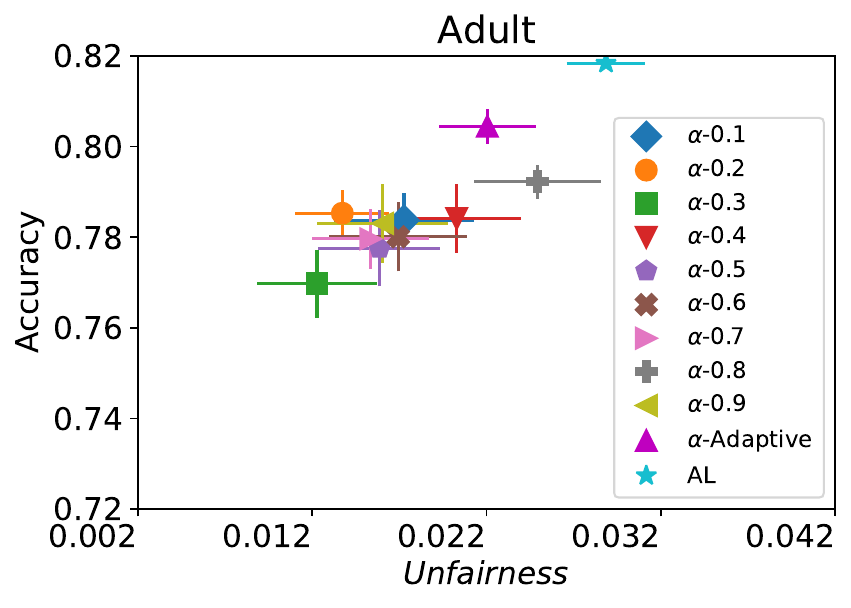}}
\subfigure[FAL Nested]{\label{exp:nested-adult}\includegraphics[width=0.32\linewidth]{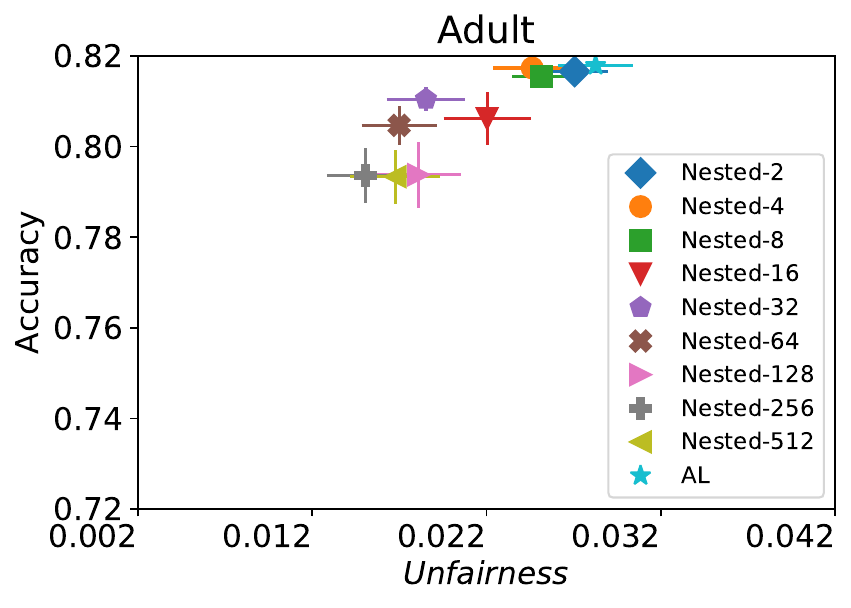}}
\subfigure[FAL Nested-Append]{\label{exp:append-adult}\includegraphics[width=0.32\linewidth]{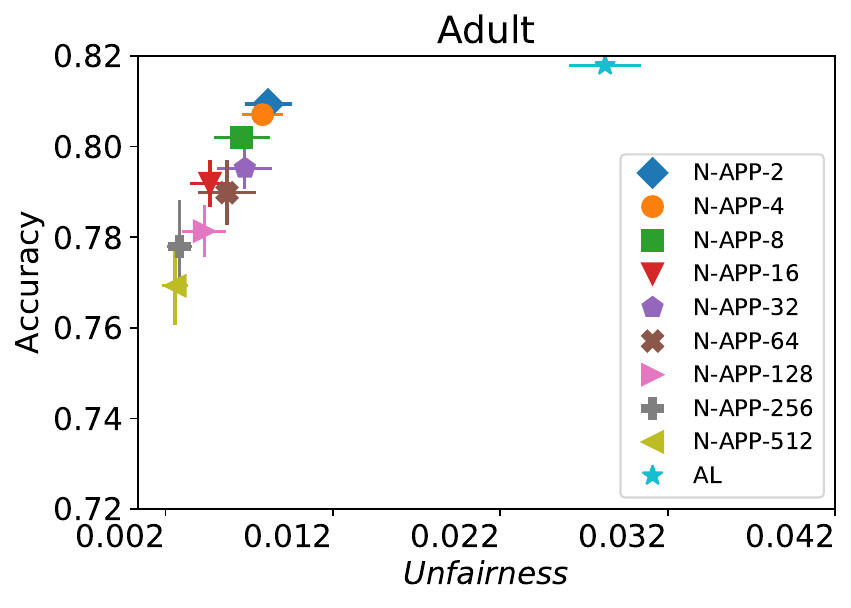}}
\vspace{-3mm}
\caption{The average DP and accuracy of different FAL approaches.}\label{exp:fal}
\end{figure*}

\begin{figure*}[!t]
\centering     %%% not \center
\subfigure[FBC $\alpha$-aggregate]{\label{exp:fbc-compas}\includegraphics[width=0.32\linewidth]{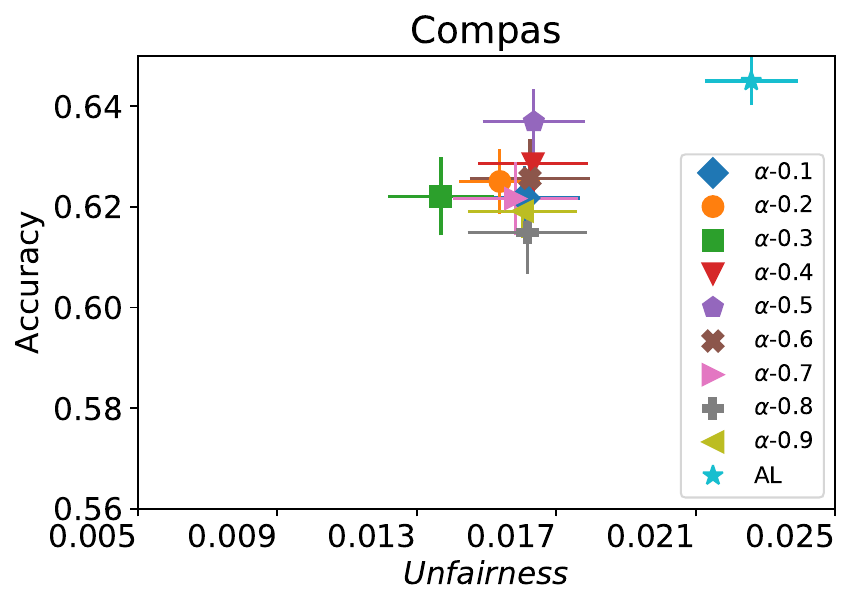}}
\subfigure[FBC Nested]{\label{exp:fbc-nested-compas}\includegraphics[width=0.32\linewidth]{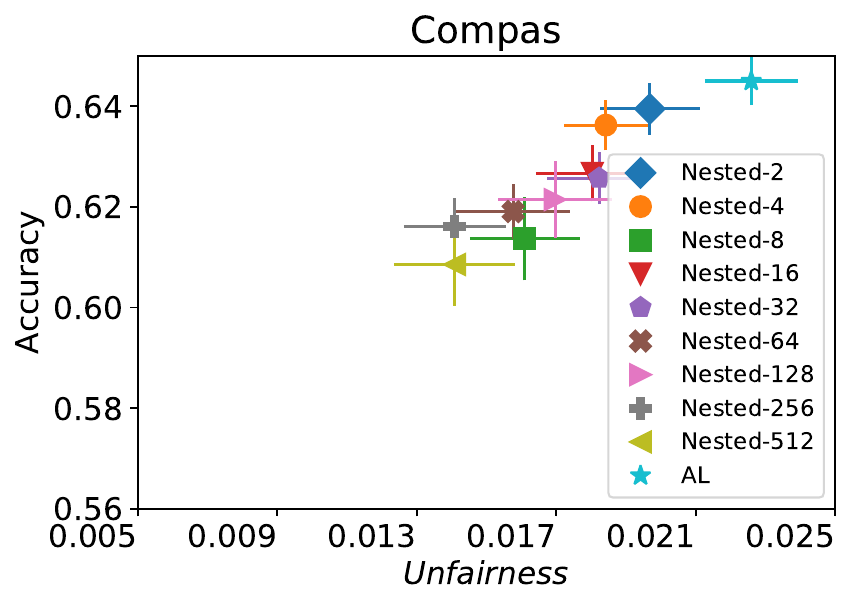}}
\subfigure[FBC Nested-Append]{\label{exp:fbc-append-compas}\includegraphics[width=0.32\linewidth]{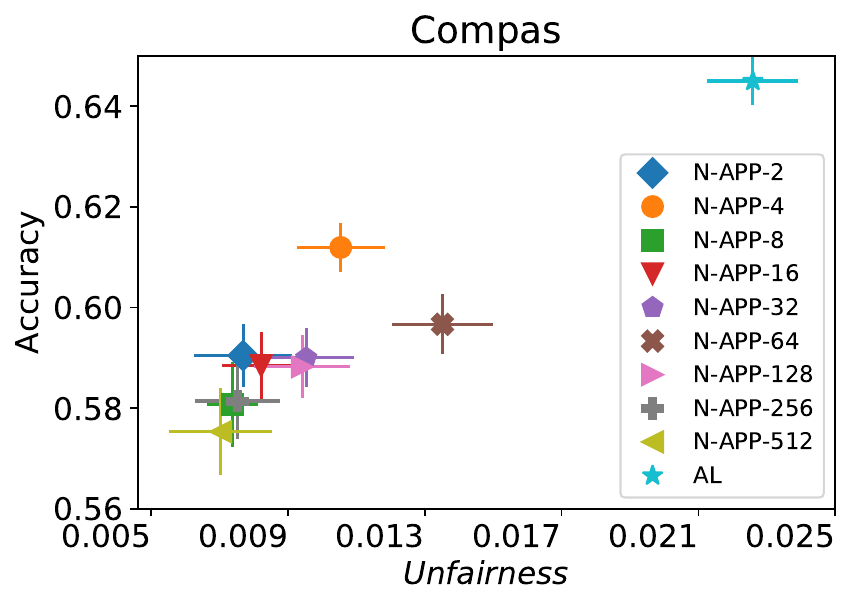}}
\vspace{-3mm}
\caption{The average DP and accuracy of different FBC approaches, COMPAS Dataset. }\label{exp:fbc}
\end{figure*}

\begin{figure*}[!ht] 
    \begin{minipage}[t]{0.66\linewidth}
        	\centering
        	\includegraphics[width=0.49\linewidth]{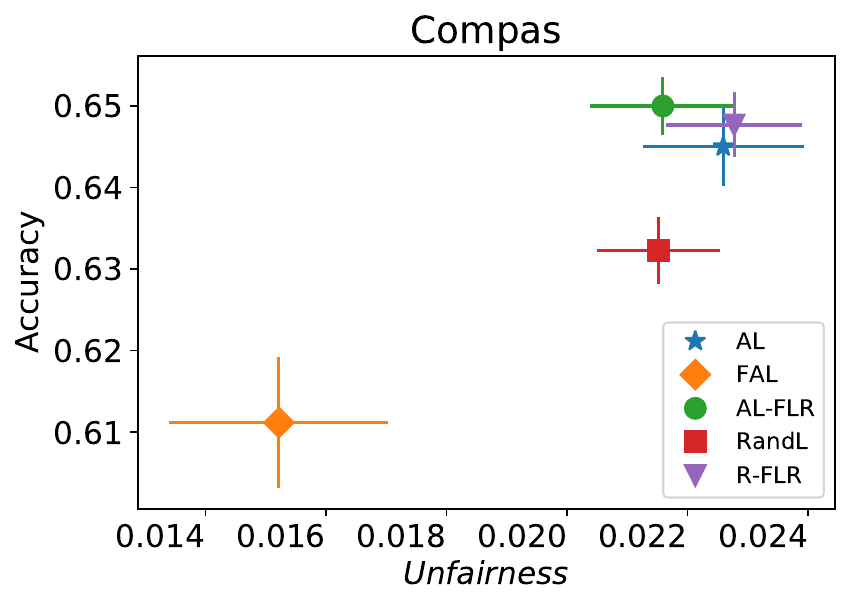}
            \includegraphics[width=0.49\linewidth]{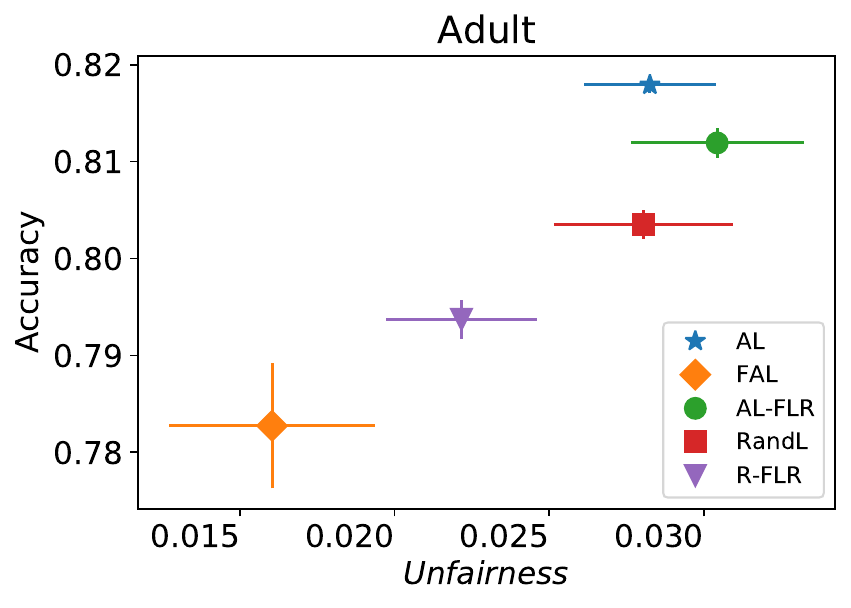}
        	\vspace{-4mm}
        	\caption{Comparison with baselines.}
            \label{exp:base}
    \end{minipage}
    \hfill
    \begin{minipage}[t]{0.32\linewidth}
        	\centering
        	\includegraphics[width =\textwidth]{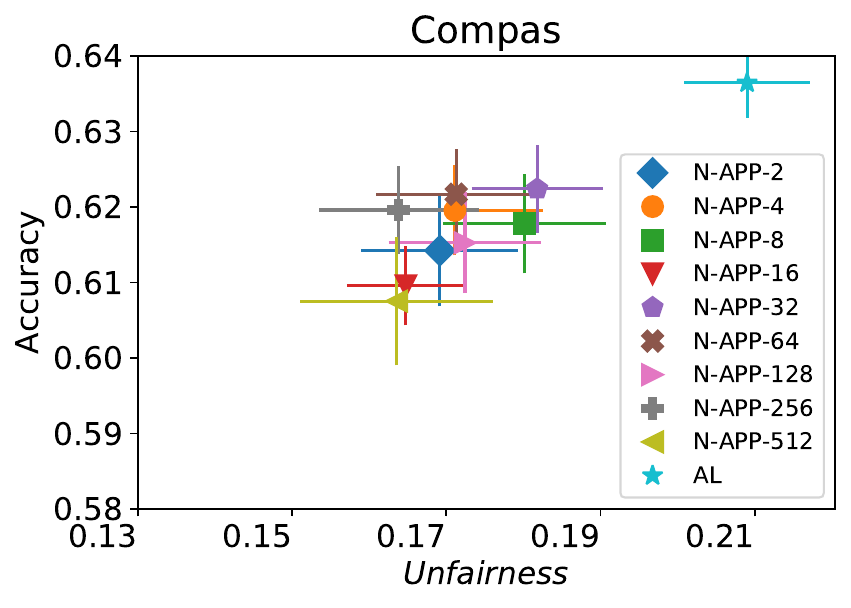}
        	\vspace{-8mm}\caption{FAL with Equalized odds, COMPAS Dataset.}
            \label{exp:odds-compas}
    \end{minipage}
    % \vspace{-4mm}
\end{figure*}

% \begin{figure*}[!t]
% \centering     %%% not \center
% \subfigure[Accuracy]{\label{exp:acc-budget}\includegraphics[width=0.32\linewidth]{figs/acc-budget.pdf}}
% \subfigure[Unfairness]{\label{exp:fair-budget}\includegraphics[width=0.32\linewidth]{figs/fair-budget.pdf}}
% \subfigure[Computation Time]{\label{exp:time}\includegraphics[width=0.32\linewidth]{figs/time.pdf}}
% \vspace{-3mm}
% \caption{--}\label{exp:budget}
% \end{figure*}

\begin{figure*}[!ht] 
    \begin{minipage}[t]{0.66\linewidth}
        \centering
        % \vspace{-39mm}
       \includegraphics[width=0.49\linewidth]{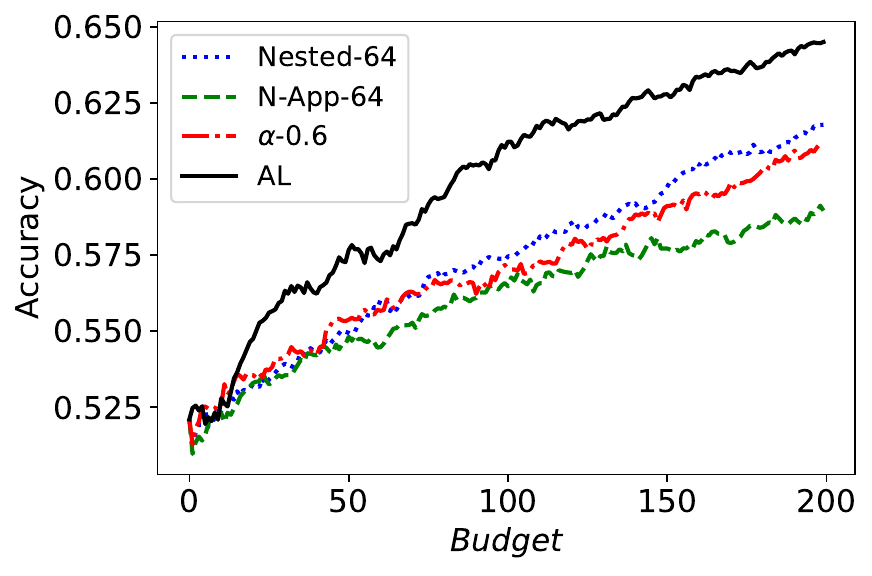}
        \includegraphics[width=0.49\linewidth]{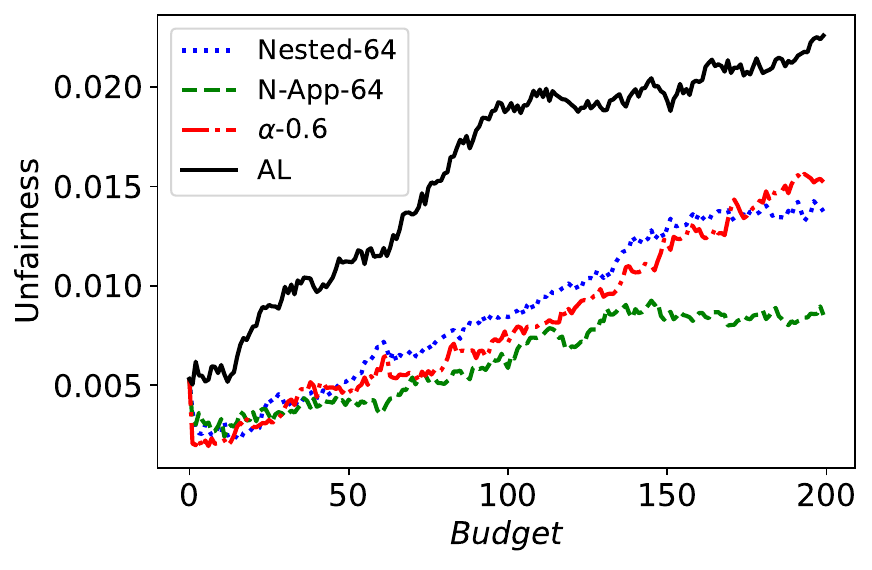}
        \vspace{-4mm}
        \caption{Performance evaluation over budget.}
        \label{exp:budget}
    \end{minipage}
    \hfill
    \begin{minipage}[t]{0.32\linewidth}
        	\centering
        	\includegraphics[width = \textwidth]{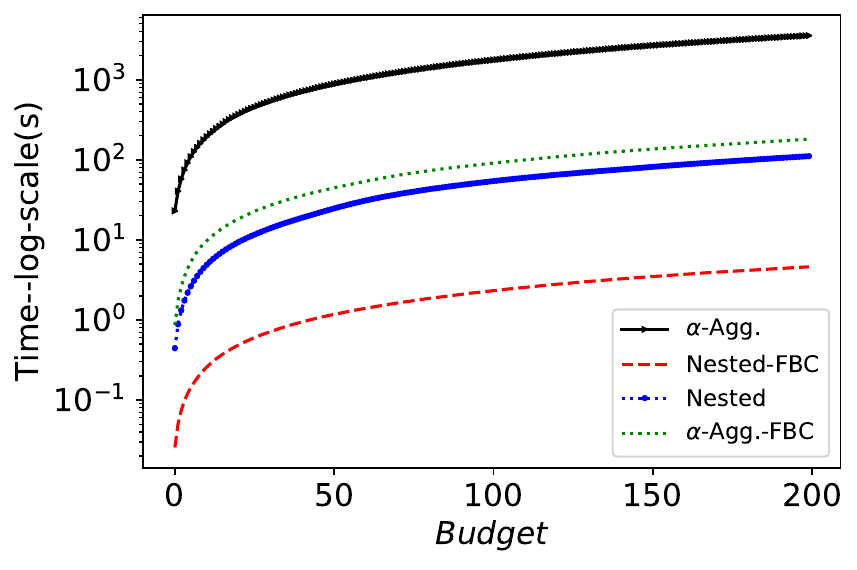}
        	\vspace{-8mm}\caption{Computation time.}
            \label{exp:time}
    \end{minipage}
    \vspace{-4mm}
\end{figure*}

%% file: conclusion.tex
\vspace{-3mm}
\section{Final Remarks}\label{sec:con}
Prior works on fair classification assume the availability of sufficiently labeled data.
In a number of societal applications such as recidivism prediction,
the labeled data is unavailable and collecting it is expensive and time-consuming.
The traditional active learning approach focuses on accuracy, often at the cost of fairness.
In this paper, we proposed a framework for fairness in active learning
that balances fairness and accuracy by selecting samples from the unlabeled pool that maximizes
a linear combination of misclassification error reduction and improvement over expected fairness.
We described a wide variety of optimizations for improving accuracy, fairness, and running time. 
Specifically, {\bf FAL Nested-Append} successfully achieves a deft balance between accuracy and fairness. 
Our extensive experiments on real datasets confirm that our proposed approach
produces a fairer model without significantly sacrificing the accuracy.
We hope that our proposed approach will have a positive impact by improving
1the model fairness in a number of real-world scenarios.

%% file: tech/fbc-maintain.tex
\section{Efficiently Computing Covariance of $X$ and $y$ in $\mathcal{L}$}\label{app:FBC-m}
%So far in this section, we proposed the efficient FAL by covariance method that works based on Equation~\ref{eq:efbc}.
When the number of features $d$ is a constant, it is possible to further improve the efficiency by
computing $E\big[FbC^{\downarrow}_{j} \big]$ in {\em constant time} by
maintaining the aggregates from the previous steps.
%Considering the number of features $d$ as a constant, we now show how, maintaining the aggregates from the previous steps, $E\big[FbC^{\downarrow}_{j} \big]$ can be computed in {\em constant time}.
Therefore, the time complexity of the fair active learning framework (without considering the labeling cost) drops from $O(B \cdot T \cdot |\mathcal{U}|)$ to $O\big(B \cdot (|\mathcal{U}|+T)\big)$, {\em the same as traditional active learning}.
%In the following we show the details of computing $E\big[FbC^{\downarrow}_{j} \big]$ for every point $P_j\in\mathcal{U}$ in constant time, maintaining the aggregates from the previous steps.

First, we note that $cov(x_i,S)$, the covariance of each feature $x_i$ with $S$, does not depend on $\mathcal{L}$ and can be computed in advance, using the unlabeled samples in $\mathcal{U}$. It is computed once for every feature at the beginning of the process and the same numbers will be used in different iterations.
The values of $cov_i = cov_{\mathcal{L}}(x_i,y)$ and $cov_{j,i,k} = cov_{\mathcal{L}\cup\{\langle X^{(j)}, S^{(j)},y^{(j)}=k \rangle\}}(x_i,y)$ in Equation~\ref{eq:efibc},
however, depend on the set of labeled data and should get recomputed at different iterations and for different points $P_j\in\mathcal{U}$.
We maintain the following aggregates for efficiently computing these values:
\begin{align*}
    &~ \mathcal{G}_y ~= \sum_{\forall \langle X^{(\ell)},S^{(\ell)},y^{(\ell)}\rangle\in\mathcal{L}} y^{(\ell)}\\
    \forall i\in[1,d]: ~& \mathcal{G}_x[i] = \sum_{\forall \langle X^{(\ell)},S^{(\ell)},y^{(\ell)}\rangle\in\mathcal{L}} X^{(\ell)}_i\\
        \forall i\in[1,d]: ~& \mathcal{G}_z[i] = \sum_{\forall \langle X^{(\ell)},S^{(\ell)},y^{(\ell)}\rangle\in\mathcal{L}} X^{(\ell)}_i y^{(\ell)}
\end{align*}

Note that at every iteration each of the above aggregates can be updated in constant time by adding the corresponding value from the new point to it.
Now, using these aggregates:
$$cov_i = cov_{\mathcal{L}}(x_i,y) = \frac{\mathcal{G}_z[i]}{n} - \frac{\mathcal{G}_x[i]}{n}\times\frac{\mathcal{G}_y}{n}$$
Similarly, for a point $P_j = \langle X^{(j)}, S^{(j)} \rangle\in\mathcal{U}$ and a label $y^{(j)}=k$:
$$cov_{j,i,k} = \frac{\mathcal{G}_z[i] + k X^{(j)}_i}{n+1} - \frac{\mathcal{G}_x[i] + X^{(j)}_i }{n+1}\times\frac{\mathcal{G}_y + k}{n+1}$$

%% file: tech/proofs.tex
\section{Proof of Lemma 1}\label{app:proof1}
% \begin{proof}
%   \textcolor{red}{S: move the proof to appendix or remove it? Reviewers mentioned it as straightforward}
\begin{align*}
    cov(S,\hat{y})&=E[S\, \hat{y}] - E[S]E[\hat{y}]\\
    E[S]E[\hat{y}] &= \mu_S E\big[\sum\theta_ix_i\big] = \mu_S \sum\theta_i\mu_{x_i}\\ &=\theta_1\mu_S\mu_{x_1}+\theta_2\mu_S\mu_{x_2}+\cdots+\theta_d\mu_S\mu_{x_d}\\
    E[S\hat{y}]&= E\big[S\,\sum\theta_ix_i\big] = E\big[\sum S\, \theta_ix_i\big] \\
    & = E\big[S\,\theta_1x_1+S\,\theta_2x_2+\cdots+ S\,\theta_d x_d \big]\\
    &=E\big[S\,\theta_1x_1\big]+E\big[S\,\theta_2x_2\big]+\cdots+ E\big[S\,\theta_d x_d \big]\\
    &=\theta_1E\big[S\,x_1\big]+\cdots+ \theta_d E\big[S\, x_d \big]
\end{align*}

\begin{align*}
    \Rightarrow cov(S,\hat{y})&=
    \theta_1E\big[S\,x_1\big]+\cdots+ \theta_d E\big[S\, x_d \big]- \\
    &~~~~(\theta_1\mu_S\mu_{x_1}+\cdots+\theta_d\mu_S\mu_{x_d})\\
    &=\theta_1(E\big[S\,x_1\big] - \mu_S\mu_{x_1})+\cdots+
    \theta_d(E\big[S\,x_d\big] - \mu_S\mu_{x_d}) \\
    &= \sum\limits_{i=1}^d \theta_i cov(S,x_i) = \theta^\top cov(S,X)
\end{align*}
% \end{proof}

%% file: plots-new-app.tex
\section{Complimentary Experiment Results}\label{app:exp}
In this section, we provide additional experimental results on the performance of our proposed algorithms, using {FBC} and Equalized Odds.

\begin{figure*}[!ht]
\centering     %%% not \center
\subfigure[FBC $\alpha$-aggregate]{\label{exp:fbc-adult}\includegraphics[width=0.32\linewidth]{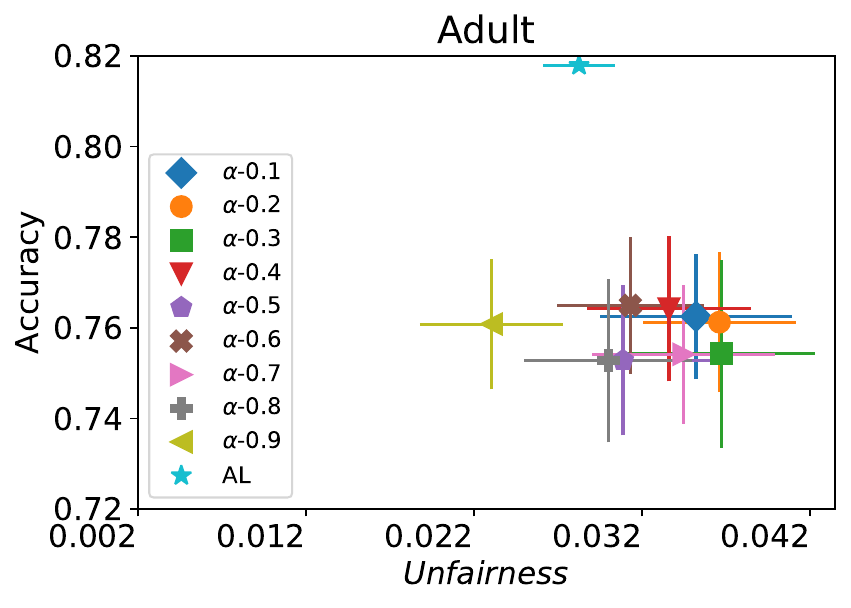}}
\subfigure[FBC Nested]{\label{exp:fbc-nested-adult}\includegraphics[width=0.32\linewidth]{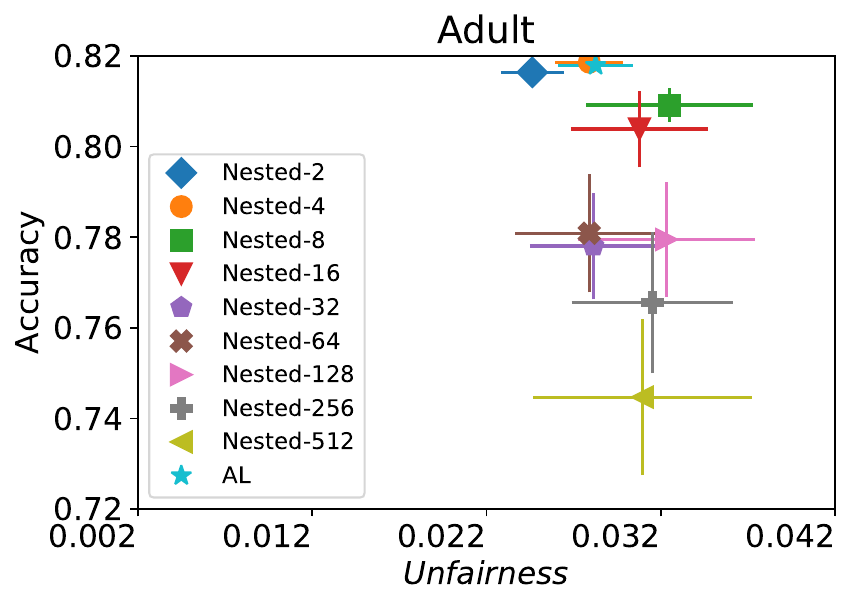}}
\subfigure[FBC Nested-Append]{\label{exp:fbc-append-adult}\includegraphics[width=0.32\linewidth]{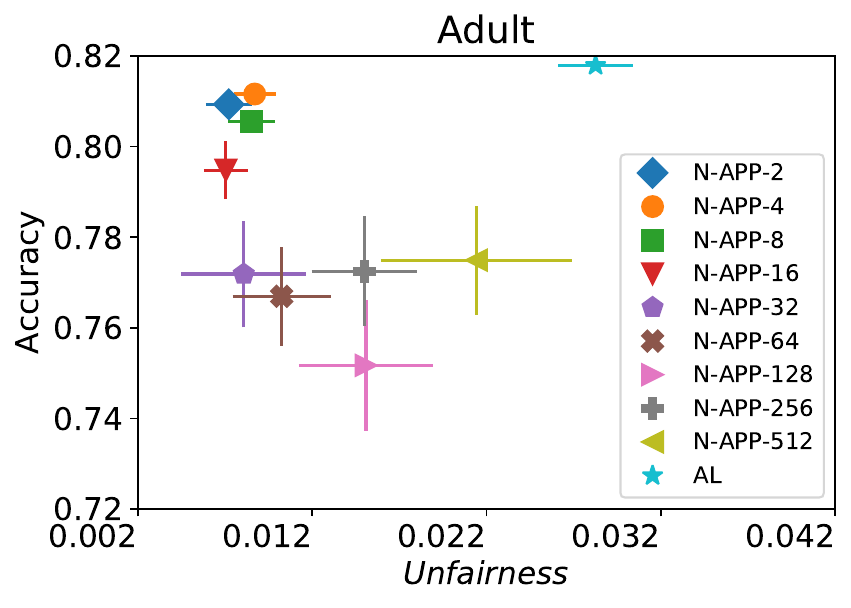}}
\vspace{-3mm}
\caption{The average DP and accuracy of different FBC approaches, Adult Dataset.}\label{exp:fbc2}
\end{figure*}

Figure~\ref{exp:fbc2} presents results for {\bf FBC} approach using three different optimizers, {\bf FAL-$\alpha$}, {\bf Nested}, and {\bf N-App} on Adult dataset. It can be seen that our ideal {\bf N-App} approach outperforms other optimizers when we use the efficient FBC approach for fairness approximation.

\begin{figure*}[!t]
\centering     %%% not \center
\subfigure[$\alpha$-aggregate]{\label{exp:odds-alpha-compas}\includegraphics[width=0.32\linewidth]{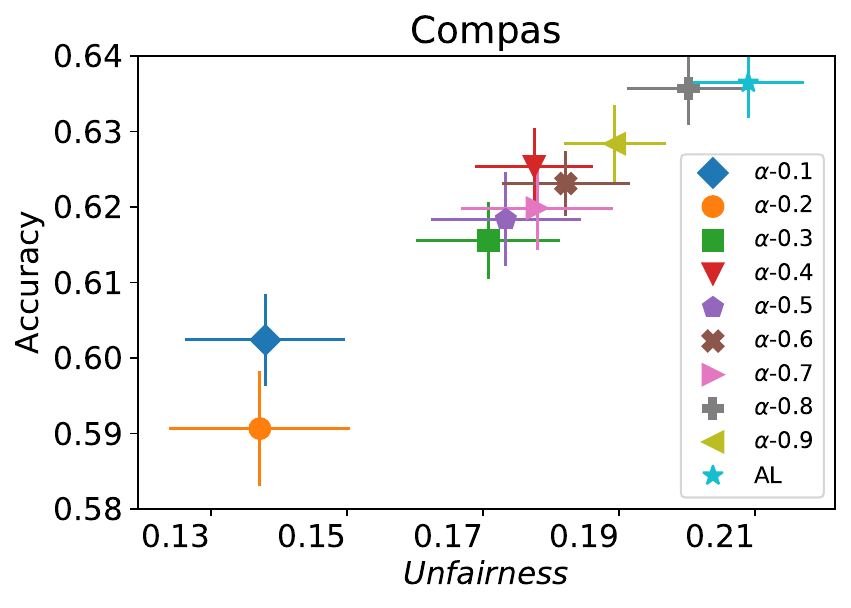}}
\subfigure[ Nested ]{\label{exp:odds-nested-compas}\includegraphics[width=0.32\linewidth]{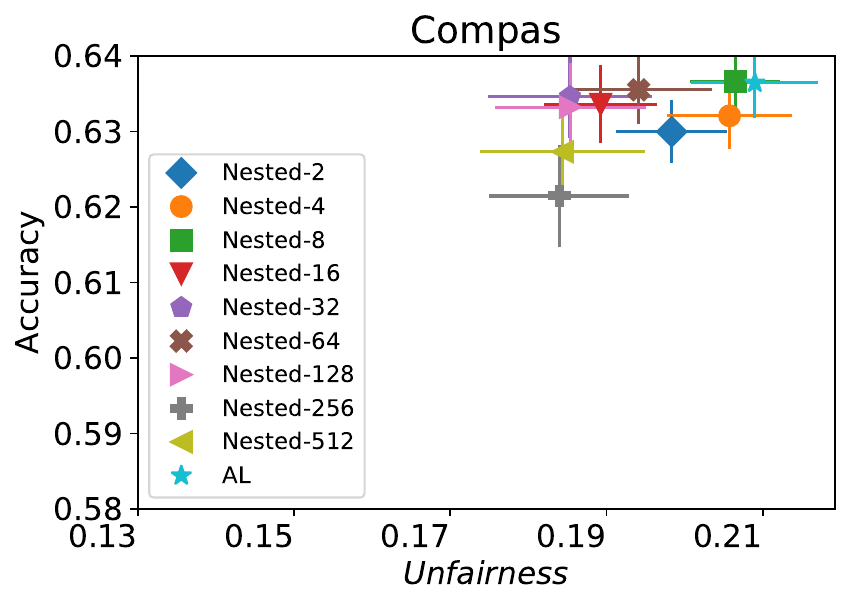}}
\vspace{-3mm}
\caption{Equalized Odds, COMPAS Dataset. }\label{exp:odds-compas-2}
\end{figure*}

In Figure~\ref{exp:odds-compas} we provided our experiment results for Equalized Odds, using {\bf N-App}, on COMPAS dataset.
Figure~\ref{exp:odds-compas-2} shows our complimentary results for the other two accuracy-fairness optimizers: {\bf FAL-$\alpha$} and {\bf Nested}.

\begin{figure*}[!t]
\centering     %%% not \center
\subfigure[$\alpha$-aggregate]{\label{exp:odds-adult}\includegraphics[width=0.32\linewidth]{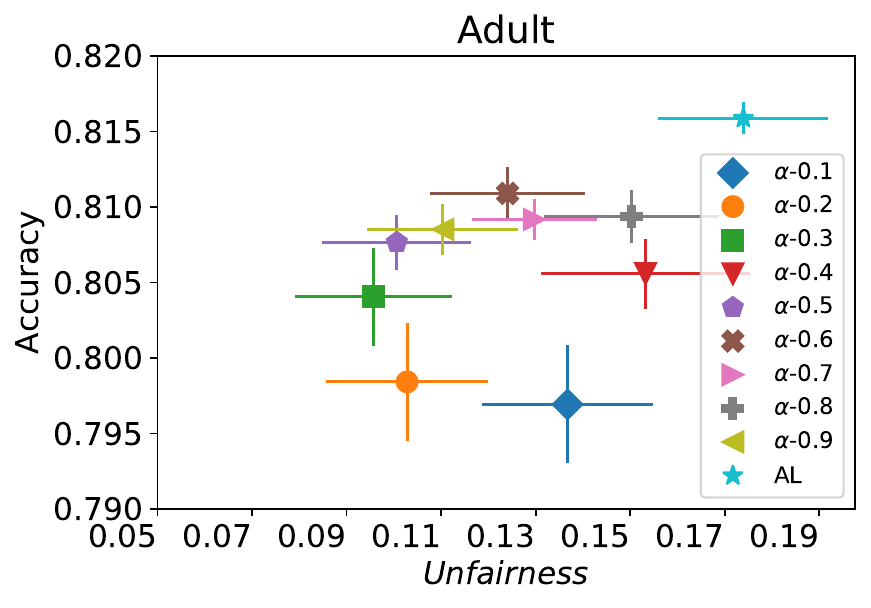}}
\subfigure[ Nested]{\label{exp:odds-nested-adult}\includegraphics[width=0.32\linewidth]{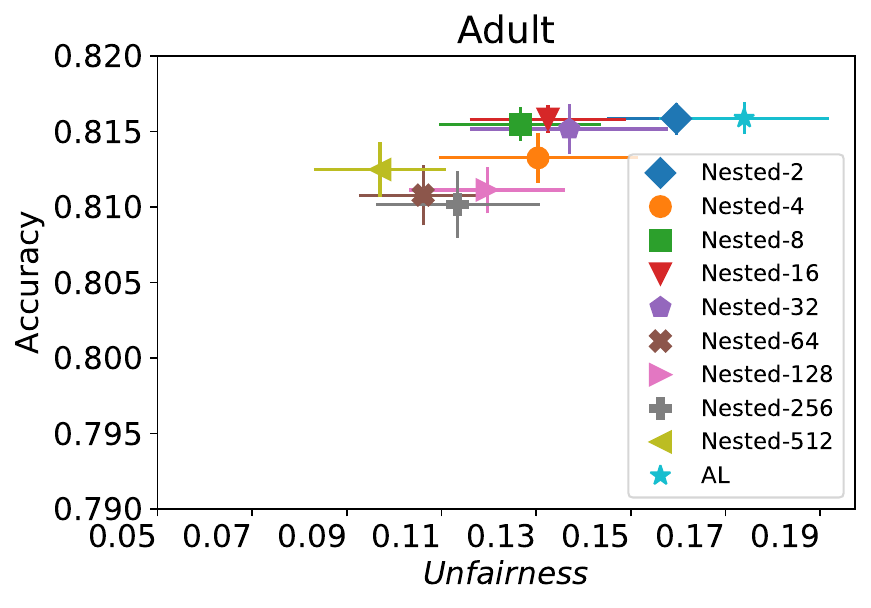}}
\subfigure[Nested-Append]{\label{exp:odds-append-adult}\includegraphics[width=0.32\linewidth]{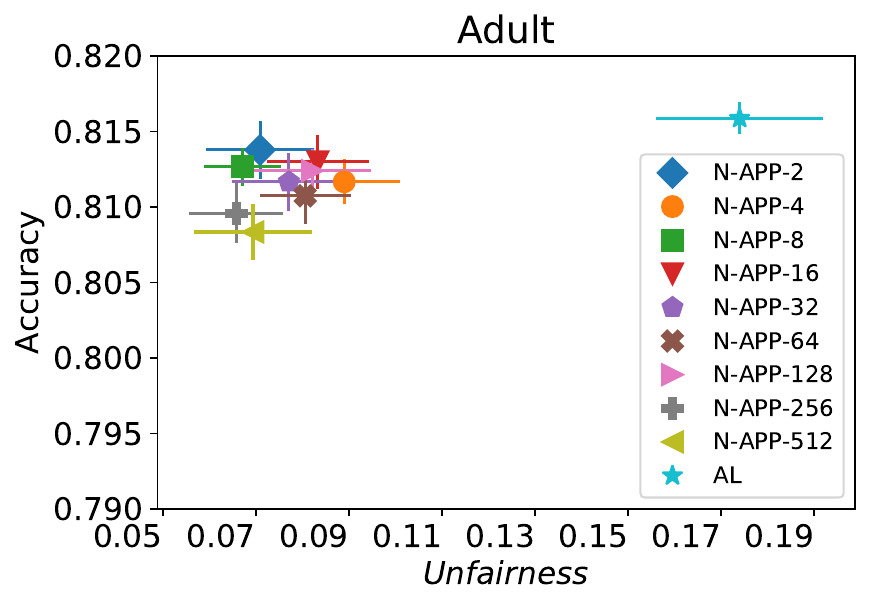}}
\vspace{-3mm}
\caption{Equalized Odds, Adult Dataset. }\label{exp:odds-adult-2}
\end{figure*}

Finally, Figure~\ref{exp:odds-adult-2} provides results of FAL using three different optimizer for Equalized Odds on Adult dataset. The results indicate that our efficient and effective {\bf N-App} approach outperforms other optimizers in terms of unfairness reduction while maintaining accuracy. 

% \begin{figure}[!ht]
% \centering     %%% not \center
% \includegraphics[width =.33\textwidth]{figs/odds-adult.pdf}
% \vspace{-3mm}
% \caption{FAL with Equalized odds, COMPAS Dataset}\label{exp:odds-adult}
% \end{figure}

